\documentclass[journal]{IEEEtran}

\usepackage{amsmath}
\usepackage{amsfonts}
\usepackage{amssymb}

\usepackage{amsthm}
\usepackage{mathtools}
\usepackage{tabularx,booktabs}
\usepackage{graphicx}
\usepackage{subfigure}
\usepackage{enumerate}
\usepackage{float}
\usepackage{url}
\usepackage{verbatim}
\usepackage{cite}
\usepackage{diagbox}
\usepackage{multirow}
\usepackage{makecell}
\usepackage{algorithm}
\usepackage{algorithmicx}
\usepackage{algpseudocode}
\usepackage{pifont}
\usepackage[linkcolor=black,citecolor=black,urlcolor=black,colorlinks=true]{hyperref}
\usepackage{bm}
\usepackage{caption}

\graphicspath{{figures/}}
\IEEEoverridecommandlockouts

\author
{
    Tiankai Yang$^{1,2}$, Kaixin Chai$^{2}$, Jialin Ji$^{1,2}$, Yuze Wu$^{1,2}$, Chao Xu$^{1,2}$, and Fei Gao$^{1,2}$
    \thanks{Corresponding author: Fei Gao}
    \thanks{\textsuperscript{1}Institute of Cyber-Systems and Control, College of Control Science and Engineering, Zhejiang University, Hangzhou 310027, China.}
    \thanks{\textsuperscript{2}Huzhou Institute, Zhejiang University, Huzhou 313000, China.}
    \thanks{E-mail: {tkyang@zju.edu.cn} {fgaoaa@zju.edu.cn} }
    \thanks{This work was supported by the national Key R\&D Program of china under grant no. 2023YFb4706600}
}

\title{\LARGE \bf Ground-Effect-Aware Modeling and Control for Multicopters}

\begin{document}
    \maketitle

\begin{abstract}

The ground effect on multicopters introduces several challenges, such as control errors caused by additional lift, oscillations that may occur during near-ground flight due to external torques, and the influence of ground airflow on models such as the rotor drag and the mixing matrix.
This article collects and analyzes the dynamics data of near-ground multicopter flight through various methods, including force measurement platforms and real-world flights. For the first time, we summarize the mathematical model of the external torque of multicopters under ground effect.
The influence of ground airflow on rotor drag and the mixing matrix is also verified through adequate experimentation and analysis. Through simplification and derivation, the differential flatness of the multicopter's dynamic model under ground effect is confirmed.
To mitigate the influence of these disturbance models on control, we propose a control method that combines dynamic inverse and disturbance models, ensuring consistent control effectiveness at both high and low altitudes.
In this method, the additional thrust and variations in rotor drag under ground effect are both considered and compensated through feedforward models.
The leveling torque of ground effect can be equivalently represented as variations in the center of gravity and the moment of inertia. In this way, the leveling torque does not explicitly appear in the dynamic model.
The final experimental results show that the method proposed in this paper reduces the control error (RMSE) by \textbf{45.3\%}.
Please check the supplementary material at:
\url{https://github.com/ZJU-FAST-Lab/Ground-effect-controller}.

\end{abstract}

\begin{IEEEkeywords}
    Ground effect, multicopter control, disturbance modeling.
\end{IEEEkeywords}

\section{Introduction}
\label{sec:intro}

\begin{figure}[t]
    \vspace{0cm}
    \begin{center}
        \includegraphics[angle=0,width=0.45\textwidth]{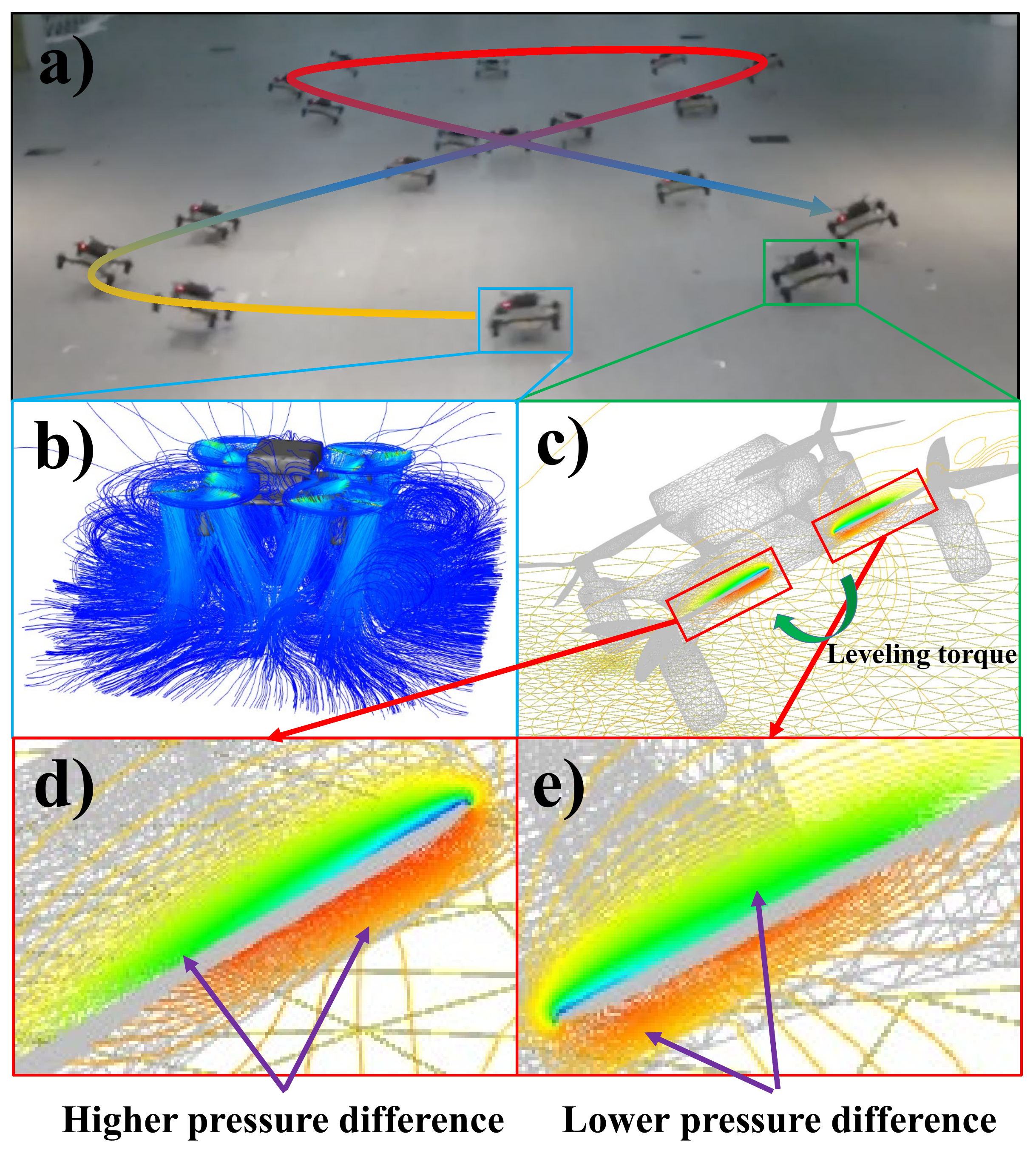}
        \caption
        {
            (a) The multicopter is flying close to the ground, ensuring a small distance from the ground surface.
            (b) The airflow around a multicopter flying near the ground. The change in the direction of the airflow results in additional lift.
            (c) Air pressure around the rotors of a tilted multicopter near the ground (red for high pressure, green for low pressure).
            (d)(e) The closer the rotor is to the ground, the greater the pressure difference between the upper and lower surfaces, resulting in a torque that tends to level the multicopter.
        }
        \label{fig_head}
    \end{center}
    \vspace{-0.9cm}
\end{figure}

\IEEEPARstart{T}hanks to the simple mechanical design and high maneuverability,
multicopters are widely used in both open sky and indoor environments.
However, near-ground operations are unavoidable in many applications.
For example, using a multicopter with an arm to grab an object close to the ground\cite{fishman2021dynamic, saunders2024autonomous}, planning a near-ground trajectory to use the extra thrust provided by ground effect to save energy\cite{gao2019exploiting}, flying in the tunnel \cite{wang2022neither} or along the wall \cite{ding2022passive} automatically landing on the surface of the object\cite{luo2015biomimetic, ji2022real}, and so on.
In these scenarios, when approaching rigid structures, multicopters are disturbed by the airflow near surfaces of ground or objects, which significantly influences the safety and stability.
Fig.~\ref{fig_head}(a) illustrates the scenario of a multicopter flying close to the ground.
In Fig.~\ref{fig_head}(b), it is evident that the airflow around the multicopter changes significantly under the influence of the ground.

This phenomenon, usually known as \textit{ground effect}, has been extensively analyzed by researchers \cite{powers2013influence, sanchez2017characterization, he2019ground, wei2019mitigating}.
Most works focus on precisely modeling the extra lift produced by ground effect,
typically with a height-dependent external force function.
In this way, a feed-forward control signal can be calculated using the model,
and is then applied to cancel the disturbance while the multicopter flying with a static height and attitude.
However, the disturbances imposed on multicopters by ground effect go beyond this.


The first one is the external torque.
When a multicopter flies near flat ground and its attitude is not horizontal, the propellers closer to the ground generate more thrust, creating a torque that tends to level the multicopter's attitude \cite{sanchez2017characterization}. We call it the \emph{leveling torque}. This could result in oscillations or a decrease in control precision.
Fig.~\ref{fig_head}(c$\sim$e) show the pressure difference between the upper and lower surfaces of a tilting multicopter's propellers, providing some insight into the source of the leveling torque.


The second one is the decrease in rotor drag.
A blade on a rotating propeller experiences lift from the air and resistance applied to the plane of rotation.
This resistance is usually proportional to the rotor speed and the forward flight speed \cite{bristeau2009role}.
Since the total thrust during high-altitude flight is maintained near the hover throttle, the rotor drag coefficient is commonly modeled as a constant\cite{faessler2017differential, svacha2017improving}.
However, as mentioned earlier, the multicopter needs a lower throttle when flying near the ground, so it can be inferred that the rotor drag also decreases.
The actual situation may be more complicated due to the presence of near-ground airflow.
We will verify whether the rotor drag decreases as expected in the following experiments.


The third one is not usually considered is the escape of the lift airflow when flying at high speed near the ground.
Kan et al. \cite{kan2019analysis} analyze the variations in additional thrust during forward flight near the ground.
When hovering, the airflow bouncing off the ground surface creates a high-pressure layer, resulting in an additional thrust for the multicopter.
During high-speed forward flight, some near-ground airflow escapes, leading to a reduction in the additional thrust.

These problems make it difficult for multicopters to maintain a tilted attitude near the ground in order to generate horizontal acceleration.
As a result, the existing methods are only suitable for situations such as taking off, landing and near-ground hovering, posing difficulties in tracking a trajectory close to the ground.
Therefore, exploring the dynamics model of the multicopter under the ground effect is necessary.

For this purpose, we establish real-world force and torque measurement platforms (Sec.~\ref{sec:method_valid}). Real-flight experiments are conducted to collect external disturbance data in various scenarios. Based on the collected data, we develop an external torque model (Sec.~\ref{subsec:getorque}) for ground effect (first-time) and validate the impact of ground effect on rotor drag (Sec.~\ref{subsec:geRD}) and mixing matrix (Sec.~\ref{subsec:bodytorque}). We simplify the drag and thrust models and convert the torque model to the payload model (Sec.~\ref{subsec:simball}).
These simplifications and equivalences make the dynamic model of a multicopter differential flat (Sec.~\ref{sec:flatness}) under the influence of ground effects, which allows to generate feedforward control commands based on the trajectory.
We combine the model-based control method (Sec.~\ref{sec:controller}) with the model-free method to effectively improve the control precision (Sec.~\ref{sec:experiment}).



This paper has the following contributions:

\begin{itemize}
    \item
    A series of methods (including force measuring platforms and real flight) are used to collect dynamic data of the multicopter under ground effect.
    \item
    For the first time, propose a model for the leveling torque. Validate other models under ground effect including rotor drag and mixing matrix.
    \item
    Simplify the dynamic model of the multicopter under ground effect to maintain differential flatness, and design a control method that takes both ground effect model and unknown disturbances into account.
\end{itemize}

\section{Related Work}

\subsection{Additional Thrust under Ground Effect}


The study of the ground effect of the blades is initially carried out on helicopters\cite{cheeseman1955effect}.
The influence of the helicopter's body shape on the parameters of the ground effect
is mainly reflected in the radius of the propeller and the distance between the blade plane and the ground.
Other influencing factors include air density, rotor speed, blade pitch angle
and the incoming flow speed in the direction of the rotation axis.

Since the helicopter has only one rotor, its dynamic model is relatively simple compared to the multicopter.
For a multicopter, an air convection area is formed under the entire body,
and its ground effect is greater than the superposition of individual rotors.
Therefore some ground effect models~\cite{sanchez2017characterization, he2019ground, kan2019analysis} are proposed to adapt to the multicopter.
Sanchez-Cuevas's work \cite{sanchez2017characterization}
introduced some correction terms from the helicopter' ground-effect model to the multicopter,
making it possible to obtain the model without identifying a series of polynomial coefficients.

\subsection{Control Method for Ground Effect}

Sanchez-Cuevas \cite{sanchez2017characterization} et al. provide the controller with attitude and thrust feedforward based on environmental information and a pre-identified model, enabling it to compensate for the ground effect caused by a "table".
Amit K Tripathi et al. \cite{TRIPATHI20173680} develop an INDI (Incremental Nonlinear Dynamic Inversion ) and a L1 adaptive controller to handle disturbances during landing, such as ground effects and unknown wind, which are compensated as modelless disturbances.
Du et al. \cite{du2016advanced} employ a similar model-free NDI approach to handle the takeoff process of multicopters.
Peng Wei et al. \cite{wei2019mitigating} model the addition thrust of  ground effect and compensate it with a model-based method (LQR feedback with attitude feedforward).


Some data-driven approaches\cite{YuCompensating, SAAT202574, KAIDataDriven, shi2019neural} rely on collecting large datasets in advance to train networks that predict disturbances based on environmental and state information. 
These predictions are then used for feedforward compensation, often achieving good control performance and adapting well to complex airflow conditions. 
However, such predictive networks typically lack an explicit mathematical representation. 
Our goal is to derive an accurate mathematical model through fundamental analysis, providing a solid foundation for future research in this area.

\subsection{Commonly Overlooked Leveling Torque Model}

The main focus of this article is to model and compensate for the leveling torque.
There are two common solutions in previous works.

The first solution is inner loop tuning.
At a certain height, when the multicopter tilts more, it experiences a stronger leveling torque from the ground effect, which can be considered linear at a small angle.
Since the leveling torque is aiming to level the multicopter's attitude, the relationship is equivalent to fixing a payload under the multicopter (described in detail in Sec.~\ref{subsec:simball}).
Furthermore, it can be equivalent to shifting the center of gravity down and increasing the moment of inertia, which can be handled by adjusting the gain of the attitude and angular velocity loops. Since the intensity of the ground effect varies at different altitudes, there should be a set of suitable parameters at each altitude. But the lack of models makes tuning parameters a trouble.
Some studies have skillfully bypassed this issue by tuning the inner loop parameters, though the adjustments are made for a specific altitude.



For example, Sanchez et al. \cite{sanchez2017characterization} concluded that the leveling torque is not a dominant factor and primarily contributes to the multicopter's stability, so it was not modeled or integrated into their control framework. Their study emphasizes the additional thrust generated by ground effect on specific propellers when the multicopter passes over obstacles, resulting in what they termed an "obstacle torque" that pushes the multicopter away from the obstacle.
Since his "obstacle torque" experiments were conducted at a specific height, the impact of the “leveling torque” can be addressed through flight control parameter tuning.


Another example is \emph{Neural Landing} \cite{shi2019neural} developed by Shi et al., where a network is designed based on multicopter dynamics and state information to predict external forces near the ground. In their approach, the leveling torque is considered less significant and not explicitly compensated for. However, their controller framework includes parameters, such as the angular velocity error gain, that help mitigate oscillations caused by the leveling torque. While these adjustments are optimized for low-altitude scenarios and landing, they may reduce control accuracy at higher altitudes, but this does not negatively impact performance during landing or near-ground flight.



The second solution is a disturbance observer.
The leveling torque can be obtained by an observer \cite{he2017modeling, he2019ground, du2016advanced,tal2020accurate} and then compensated.
Such methods are not effective enough.
They believe that the external torque in the next time period is about equal to the one observed in the previous time period.
It is effective for some slowly changing external disturbances.
For the shift of the center of gravity in the \emph{Z axis} or the ground effect, the external torque is proportional to the tilt angle and changes at high frequency, resulting in a delay in the disturbance estimation, and the compensation cannot eliminate the oscillations.




\begin{figure}[t]
    \vspace{0cm}
    \begin{center}
        \includegraphics[angle=0,width=0.4\textwidth]{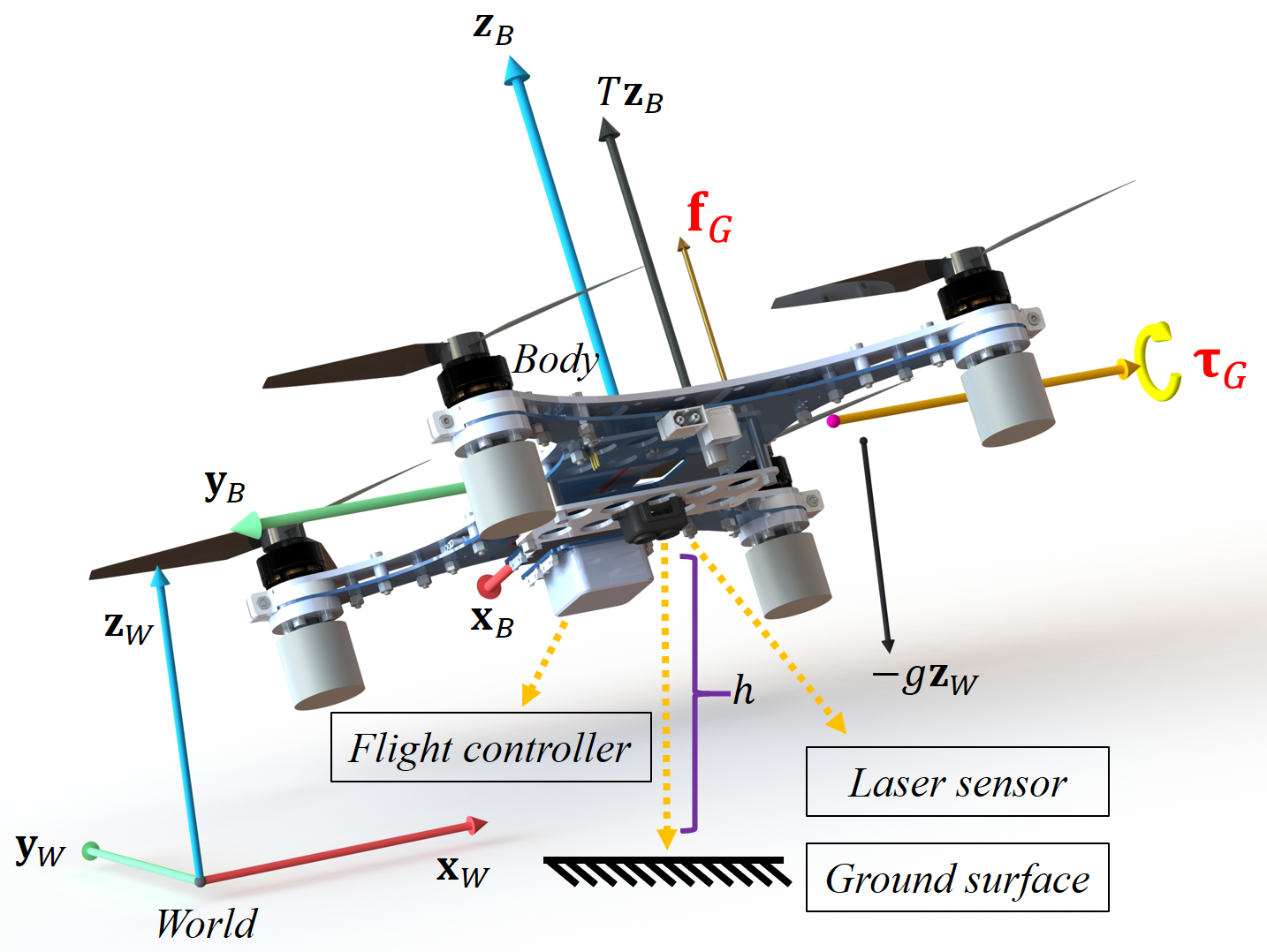}
        \caption{The coordinate systems of the multicopter.}
        \label{fig_frame}
    \end{center}
    \vspace{-0.7cm}
\end{figure}

\section{Basic Symbol Representation}
\label{sec:symbol}

Some coordinate systems, forces, and torques in this work are shown in Fig.~\ref{fig_frame}.
We use
$ \left[ {{{\bm{x}}_W},{{\bm{y}}_W},{{\bm{z}}_W}} \right] = {{\bm{I}}_3} $
to represent the 3 orthonormal bases of the \emph{world coordinate system}.
${{\bm{I}}_3}$ is the identity matrix.
The oriention of the multicopter is denoted by
${\bm{R}} = \left[ {{{\bm{x}}_{\bm{B}}},{{\bm{y}}_{\bm{B}}},{{\bm{z}}_{\bm{B}}}} \right]$,
where $\left\{ {{{\bm{x}}_{\bm{B}}},{{\bm{y}}_{\bm{B}}},{{\bm{z}}_{\bm{B}}}} \right\}$
are a set of orthogonal bases of the \emph{body coordinate system} of the multicopter.

The position of the multicopter's geometric center in the \emph{world coordinate system} is $\bm{p}$,
and $\left\{ {{\bm{v}},{\bm{a}},{\bm{j}}} \right\}$
are the velocity, acceleration and jerk of the multicopter in the \emph{world coordinate system} respectively.
The Euler angles of the multicopter in the \emph{world coordinate system}
are expressed in the order of $Z - Y - X$ rotation as
${\bm{\xi }} = \left\{ {\varphi ,\gamma ,\phi } \right\}$.
The rotation matrix around an axis of $Z - Y - X$ can be expressed as
$ {T_Z}\left( \varphi  \right) $,
$ {T_Y}\left( \gamma  \right) $ and
$ {T_X}\left( \phi  \right) $ respectively.
The angular velocity and angular acceleration of the multicopter in the \emph{body coordinate system} are expressed as
${\bm{\omega }}$ and ${\bm{\beta }}$.

The dynamics model of the multicopter can be written in the following form:
\begin{equation}
\begin{aligned}
\label{equ_euler}
    m{\bm{a}} & =  - g{{\bm{z}}_W} + T{{\bm{z}}_B} + {{\bm{f}}_G} + {{\bm{f}}_D}\\
    {\bm{J\dot \omega }} & =  - {\bm{\omega }} \times {\bm{J\omega }} + {{\bm{\tau }}_B} + {{\bm{\tau }}_G} + {{\bm{\tau }}_{ext}},
\end{aligned}
\end{equation}
where $m$ and ${\bm{J}}$ are the mass and the inertia moment of the multicopter respectively,
${\bm{g}}$ is the acceleration of gravity, 
${T{{\bm{z}}_B}}$ is the thrust from rotors,
${{{\bm{f}}_G}}$ the additional force,
${{{\bm{\tau }}_B}}$ is the torque from rotors,
${{{\bm{f}}_D}}$ is the rotor drag,
${{{\bm{\tau }}_G}}$ is the leveling torque
${{{\bm{\tau }}_{ext}}}$ is the unmodeled external torques.



For a quadrotor, the total thrust and torque in the body coordinate system generated by the rotors are:
\begin{equation}
\begin{aligned}
\label{equ_mixmatrix}
    \left[
        {\begin{array}{*{20}{c}}
        T\\
        {{{\bm{\tau }}_B}}
        \end{array}}
    \right]
    = {\bm{M}}{{\bm{N}}^2}
    &   = diag\left( {{k_T},\frac{{\sqrt 2 b}}{4}{k_{TX}},\frac{{\sqrt 2 b}}{4}{k_{TY}},{k_I}} \right) \\
    &   \left[
            {\begin{array}{*{20}{c}}
            1&1&1&1\\
            { - 1}&1&1&{ - 1}\\
            { - 1}&1&{ - 1}&1\\
            { - 1}&{ - 1}&1&1
            \end{array}}
        \right]
    \left[
        {\begin{array}{*{20}{c}}
        {{n_1}^2}\\
        {{n_2}^2}\\
        {{n_3}^2}\\
        {{n_4}^2}
        \end{array}}
    \right].
\end{aligned}
\end{equation}
$b$ represents the distance between the diagonal rotors.
$ n_i $ represents the rotation speed of each rotor.
$ {{k_T}} $ is the thrust coefficient.
$k_{TX}$ and $k_{TY}$ are the torque coefficients in the roll and pitch directions, respectively. $k_{I}$ is the inverse torque coefficient of the rotors.
$ \bm{M} $ is the mixing matirx of the quadrotor.
${n_i}$ is the rotor speed (unit: rotation-per-minute, $\rm{rpm}$),
${{\bm{N}}^2}$ is a $4 \times 1$ matrix consisting of the square of the rotor speeds.

The mechanical frame of the quadrotor is not completely centrosymmetry, and the body obstructs the airflow of the rotor, therefore it can be assumed that
$ {k_T} \ne {k_{TX}} \ne {k_{TY}} $.


\begin{figure}[ht]
    \vspace{0cm}
    \begin{center}
        \includegraphics[angle=0,width=0.48\textwidth]{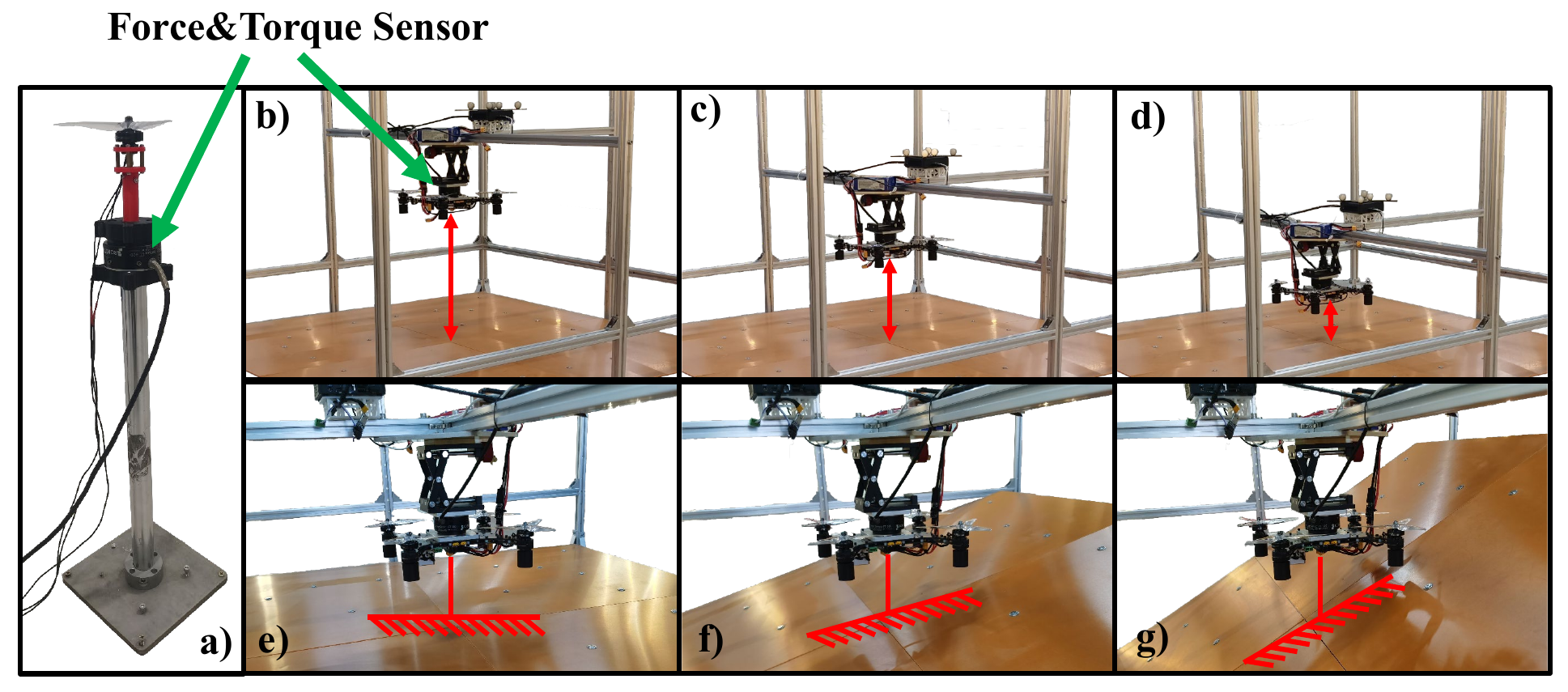}
        \caption
        {
            The single rotor (a) and quadrotor (b$\sim$g) platform.
        }
        \label{fig_model_valid}
    \end{center}
    \vspace{-0.5cm}
\end{figure}

\section{Methods for Exploring and Validating Models}
\label{sec:method_valid}


In this section, we will introduce how to set up the environments to collect and analyze the dynamic data of multicopters under ground effect.
The environments include
the single-rotor platfrom (Fig.~\ref{fig_model_valid}a),
the quadrotor platform (Fig.~\ref{fig_model_valid}(b$\sim$g) ),
realworld flight, etc.
Additionally, we introduce the \emph{Spearman correlation coefficient}\cite{de2012jackknife} to ascertain the specific factors related to external disturbances.

\subsection{Dynamic Measurement Platform in real-world}
\label{subsec:method_realplatform}


We set up dynamics testing platforms for a single rotor and a quadrotor separately. Both of them are equipped with a six-axis force and torque sensor.

\subsubsection{The single-rotor platform}
    The platform (Fig.~\ref{fig_model_valid}a) is fixed on the ground, which can measure the rotor speed, thrust, and torque of a single motor when it rotates.
    The rotor is positioned far away from the ground to minimize ground effect. Some connecting mechanisms use soft 3D-printed parts (TPU material), effectively reducing the noise in sensor data caused by rotor rotation.

\subsubsection{The quadrotor platform}


In the platform illustrated in Fig.~\ref{fig_model_valid}(b$\sim$g), the entire quadrotor is fixed on the aluminum profile frame.
A force and torque sensor is fixed between the frame and the quadrotor.
A resin board ($1.2m \times 1.2m$) is placed under the quadrotor to simulate the ground.
A series of mechanical structures are available for adjusting the tilt angle of the board and the height of the quadrotor.
This setup allows for simulating quadrotor flights near the ground with a forward tilt angle.
An array laser sensor is used to measure the distance between the ground board and the quadrotor. The tilt angle of the ground is determined by an electronic inclinometer.


It's worth noting that we adjust the angle of the ground board rather than the angle of the quadrotor.
In other words, we simulate the quadrotor tilting by tilting the ground board while keeping the quadrotor level.
This is because if the quadrotor tilts, it will generate a gravity moment associated with the tilt angle.
Additionally, when the rotors rotate, there will be a torque related to the rotor speed since the sensor center does not coincide with the thrust line of the rotors.
Both of these factors would interfere with the measurement of the torques generated by ground effect.

\subsection{Additional Methods}
\label{subsec:method_add}

\subsubsection{Real-world flight}

In addition to using the dynamic measurement platform, we also conduct real-world flight experiments to collect disturbance data experienced by the quadrotor during near-ground flights.

We introduce the following external disturbance observer:

\begin{equation}
\begin{aligned}
\label{equ_ext_force_ob}
    \begin{array}{*{20}{l}}
        {{{{\bm{\tilde a}}}_{ext}} = {{\left( {{\bm{a}} - {\bm{g}}} \right)}_f}^{imu} - {{\bm{z}}_B}{T_f}/m}\\
        {{{{\bm{\tilde \tau }}}_{ext}} = \bm{J}{{{\bm{\dot \omega }}}_f} + {{\bm{\omega }}_f} \times {\bm{J}}{{\bm{\omega }}_f} - {{\bm{\tau }}_B}}.
    \end{array}
\end{aligned}
\end{equation}

The subscript ${\left(  \cdot  \right)_f}$ indicates that the data has been filtered through a low-pass filter.
$ {\left( {{\bm{a}} - {\bm{g}}} \right)^{imu}} $
is the raw acceleration data captured by the Inertial Measurement Unit (IMU),

\subsubsection{Mechanical model}


We can directly determine the inertia moment and the center of gravity of the quadrotor with the Mechanical model.
Since the total mass of the Mechanical model is nearly equal to the actual total mass, it can be assumed that the calculated inertia moment from the Mechanical model is also nearly equal to the actual one.
In addition to being used for model exploration and validation, the controller also directly calls these parameters in practice.

\subsubsection{Correlation coefficient}

To explore and validate the relationship between external disturbances introduced by ground effect and various variables (height, rotor speed, tilt angle, etc), we calculate the \emph{Spearman's rank correlation coefficient}~\cite{de2012jackknife}.
This statistical tool assesses the existence and strength of monotonic relationships between nonlinear parameters.

For two datasets, ${x_i}$ and ${y_i}$, of the same size $n$, both sets are ranked as $R\left( {{x_i}} \right)$ and $R\left( {{y_i}} \right)$. The Spearman's rank correlation coefficient ${r_s}$ can be computed:

\begin{equation}
\begin{aligned}
\label{equ_spearman_rank}
    {r_s}\left( {x,y} \right) = \frac{{{\rm{cov}}\left[ {R\left( x \right),R\left( y \right)} \right]}}{{\sigma \left[ {R\left( x \right)} \right]\sigma \left[ {R\left( y \right)} \right]}}.
\end{aligned}
\end{equation}
${{\mathop{\rm cov}} \left[ {R\left( x \right),R\left( y \right)} \right]}$ represents the covariance of the ranked variables, while ${\sigma \left[ {R\left( x \right)} \right]}$ and ${\sigma \left[ {R\left( y \right)} \right]}$ signify their respective standard deviations.


The variables we focus on are height, tilt angle, and rotor speed.
When exploring the correlation between external disturbances and rotor speeds, it is difficult to establish a direct connection between the speed of individual rotors and the overall dynamics of the quadrotor. Consequently, the rotor speeds are transformed into ${\bm{N}_{base}}$:

\begin{equation}
\begin{aligned}
\label{equ_Nbase}
    {\bm{N}_{base}} = {\rm{diag}}{\left( {{k_T},{k_{TX}},{k_{TY}},{k_I}} \right)^{ - 1}}{\bm{M}}{{\bm{N}}^2}.
\end{aligned}
\end{equation}

In data analysis,
${\bm{N}_{base}}\left( 1 \right)$ represents the composite rotor speeds corresponding to total thrust $T$,
while ${\bm{N}_{base}}\left( 2 \right)$, ${\bm{N}_{base}}\left( 3 \right)$,
and ${\bm{N}_{base}}\left( 4 \right)$
represent the composite rotor speeds corresponding to roll, pitch, and yaw torque
(
    ${{\bm{x}}_W}^ \top {{\bm{\tau }}_B}$,
    ${{\bm{y}}_W}^ \top {{\bm{\tau }}_B}$,
    ${{\bm{z}}_W}^ \top {{\bm{\tau }}_B}$
),
respectively.



Spearman's rank correlation coefficient between disturbances and variables and the results of parameter identification are shown in MAT.Part.4\cite{groundeffectgithub}.
Relevant experiments are described in Sec.~\ref{sec:gemodel}.


\begin{figure}[ht]
    \vspace{0cm}
    \begin{center}
        \includegraphics[angle=0,width=0.4\textwidth]{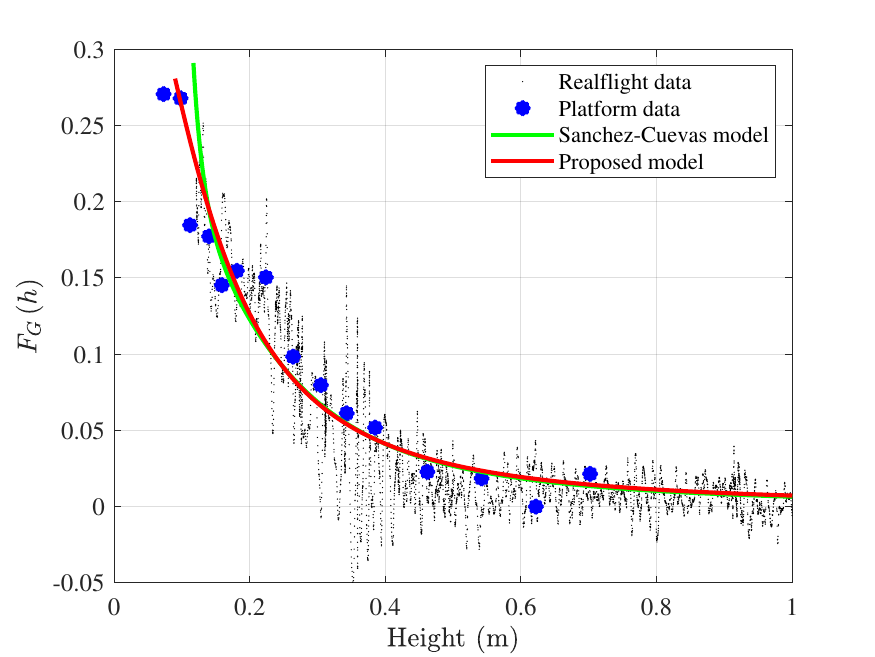}
        \caption{Data and model of the additional thrust $ {{\bm{f}}_G} $.}
        \label{ge_thrust_realworld_flight}
    \end{center}
    \vspace{-0.5cm}
\end{figure}

\section{Models under Ground effect}
\label{sec:gemodel}

\subsection{Thrust Model under Ground effect}
\label{subsec:rotors_force}

\subsubsection{Model}

The total thrust generated by the rotors is:

    \begin{equation}
    \begin{aligned}
    \label{equ_rpm_to_thrust}
        T = \sum\limits_{i = 1}^4 {{k_T}{n_i}^2}.
    \end{aligned}
    \end{equation}

The additional force generated by ground effect is:
\begin{equation}
\begin{aligned}
\label{equ_geforce}
    {{\bm{f}}_{\bm{G}}} = {F_G}\left( h \right)T{{\bm{z}}_B}.
\end{aligned}
\end{equation}

${F_G}\left( h \right)$ is a function related to the distance $h$ from the ground to the center of the rotor plane:

\begin{equation}
\begin{aligned}
\label{equ_fgh}
    {F_G}\left( h \right) = \frac{{{g_2}}}{{{h^2} + {g_1}}},
\end{aligned}
\end{equation}
where $g_1$ and $g_2$ are constants.


In previous studies\cite{sanchez2017characterization, he2019ground, kan2019analysis}, various forms of the function $ {F_G}\left( h \right) $ have been employed.
We choose the above form (\ref{equ_fgh}) primarily because of its simplicity, fewer parameters, and most importantly, its derivative form can explain the model of the leveling torque (\ref{equ_Mh}) by ground effect, which is proposed in Sec.~\ref{subsec:getorque}.

\subsubsection{Experiment}

We conduct a series of experiments to verify the model of the additional thrust model under ground effect and identify the relevant parameters.


During the quadrotor platform experiment, we align the ground board horizontally, as illustrated in Fig.~\ref{fig_model_valid}(b). We adjust the distance between the ground board and the quadrotor, collecting essential data including rotor speed and height above the ground.
The ${F_G}\left( h \right)$ function is measured using the following formula:
\begin{equation}
\begin{aligned}
\label{equ_fgh_platform}
    {\tilde F_G}\left( h \right) = \frac{{{{\bm{z}}_W}^ \top {\bm{\tilde f}}}}{{\tilde T}} - 1.
\end{aligned}
\end{equation}
${{\bm{\tilde f}}}$ and $ {\tilde T} $ are the measurement of thrust measured by the force sensor and rotor speeds (\ref{equ_rpm_to_thrust}) respectively.


During the real-world flight experiment, the quadrotor gradually reduces its hover altitude,
and the relevant data (IMU and rotor speed) is collected for estimating the ${F_G}\left( h \right)$ function:
\begin{equation}
\begin{aligned}
\label{equ_fgh_realflight}
    {{\tilde F}_G}\left( h \right) = \frac{{m{{\bm{z}}_W}^ \top {{{\bm{\tilde a}}}_{ext}}}}{{\tilde T}},
\end{aligned}
\end{equation}
where $ {{{{\bm{\tilde a}}}_{ext}}} $ is the external force obtained by (\ref{equ_ext_force_ob}).


Subsequently, the data are shown in Fig.~\ref{ge_thrust_realworld_flight}.
Our model in (\ref{equ_fgh}) matches the data from the platform experiment and the flight experiment.
It is generally consistent with Sanchez's model\cite{sanchez2017characterization}.
The associated parameters can be identified and are presented in MAT.Part.4.1 \cite{groundeffectgithub}.


To prove that ${{\bm{f}}_G}$ remains unaffected by the tilt angle of the quadrotor and is oriented in the z-axis of the quadrotor body coordinate system, we gradually adjuste the tilt angle and height of the ground board as illustrated in Fig.~\ref{fig_model_valid}(b$\sim$g) and collect relevant data..



In MAT.Part.4.2 \cite{groundeffectgithub}, correlation coefficients between ${{\bm{f}}_G}$ and various variables are computed: ${r_s}\left[ {{{\bm{f}}_G},{\bm{N}_{base}}\left( 1 \right)} \right] = 0.0054$, indicates a lack of correlation between ${{\bm{f}}_G}$ and the tilt angle $\delta $.


\subsection{The Leveling Torque Generated by Ground Effect}
\label{subsec:getorque}


During low-altitude flight, the quadrotor's tilt angle relative to the ground leads to varying ground effects on its rotors.
Rotors closer to the ground experience a more significant additional thrust, creating an external leveling torque (${{\bm{\tau }}_G}$) that tends to level the quadrotor's attitude.

\subsubsection{Model assumptions}
\label{subsubsec:assumption}

We can make some reasonable assumptions about the model of leveling torque:

$ {{\bm{\tau }}_G}$ and $ \delta $:
As the tilt angle increases, the difference in distance between the diagonal rotors and the ground increases, resulting in a larger thrust difference and an increase in leveling torque.

${{\bm{\tau }}_G}$ and $ T $:
As the total thrust increases, the additional thrust from ground effect also increases. This results in a proportional increase in thrust difference between higher and lower rotors, leading to an increase in leveling torque.

${{\bm{\tau }}_G}$ and $ h $:
If the absolute value of the derivative of the additional thrust function $ {\dot F_G}\left( h \right) $ is large at a certain height, it means that near that height, rotors at different heights will produce a larger thrust difference, resulting in a larger leveling torque.

\begin{figure}[h]
    \vspace{-0.3cm}
    \begin{center}
        \includegraphics[angle=0,width=0.45\textwidth]{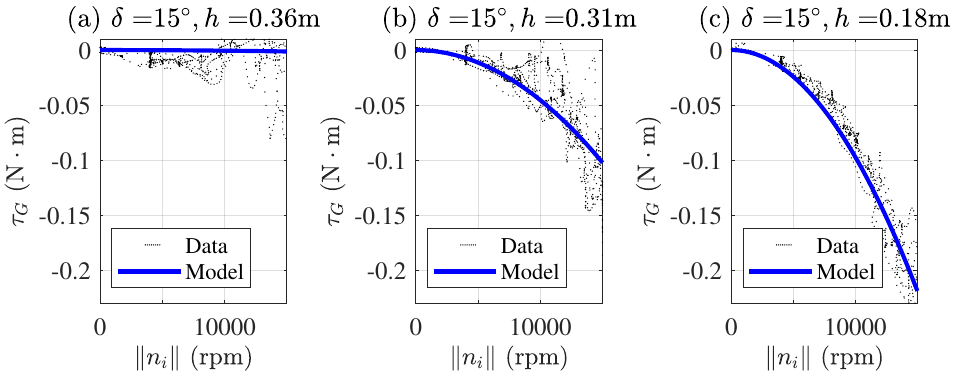}
        \caption{Relationship between leveling torque and average rotor speed. More data is in MAT.Part.2.2\cite{groundeffectgithub}.}
        \label{fit_getorque_rpm}
    \end{center}
    \vspace{-0.7cm}
\end{figure}

\begin{figure}[ht]
    \vspace{0cm}
    \begin{center}
        \includegraphics[angle=0,width=0.45\textwidth]{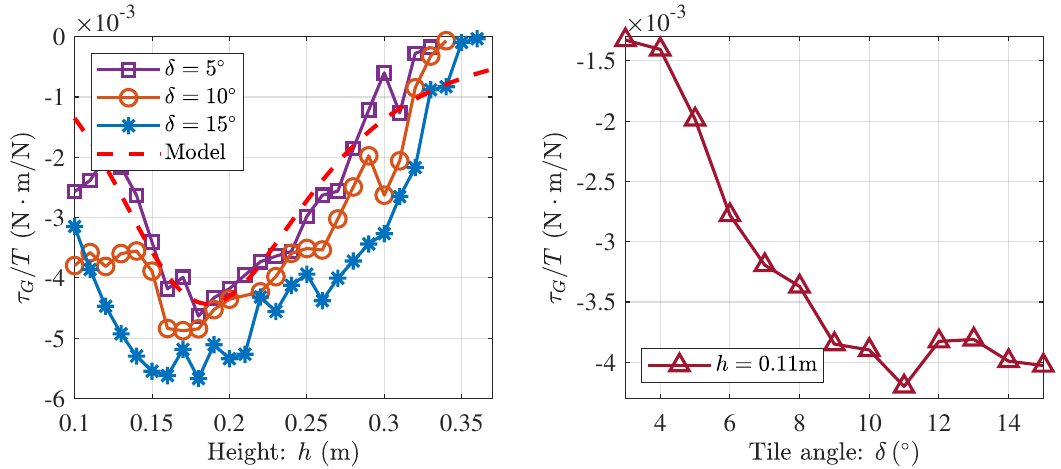}
        \caption{Leveling torque measured on the quadrotor platform. The vertical axis represents the leveling torque generated per unit of thrust. Each data point is obtained by fitting data across all rotational speeds at a specific tilt angle and altitude.}
        \label{getorque_combine}
    \end{center}
    \vspace{-0.5cm}
\end{figure}

\subsubsection{Mathematical explanations}


We conduct some mathematical derivations to explain these assumptions and simulation results.


Assuming the additional force ${{\bm{\tau }}_G}$ by ground effect is distributed on a circle with the quadrotor's center as its center and the diagonal wheelbase $b$ as its diameter.
The center of this circle is located at a height $h$ above the ground, and the height of the points on the circle is $H\left( \theta  \right)\left\{ {\theta  \in \left[ {0,2\pi } \right)} \right\}$.
The point closest to the ground corresponds to $\theta {\rm{ = }}0$.
${D\left( \theta  \right)}$ represents the distribution density function of the ground effect force ${{\bm{\tau }}_{\bm{G}}}$ on the circle. The relationship can be expressed as follows:
\begin{equation}
\begin{aligned}
\label{equ_getorque_model}
    \int_0^{2\pi } {D\left( \theta  \right)d\theta = {F_G}\left( h \right)T} \\
    D\left( \theta  \right) = {F_G}\left[ {H\left( \theta  \right)} \right]T/2\pi \\
    H\left( \theta  \right) = h - \frac{b}{2}\sin \delta \cos \theta ,
\end{aligned}
\end{equation}
where $\delta$ is the angle between the Z-axis of the body system and the world coordinate system.

We assume that around a certain height, $ {\dot F_G}\left( h \right) $ is a constant value:
\begin{equation}
\begin{aligned}
\label{equ_Fgh_diff}
    {F_G}\left[ {H\left( \theta  \right)} \right]{\rm{ = }}{F_G}\left( h \right) - \frac{b}{2}\sin \delta \cos \theta {\dot F_G} \left( h \right).
\end{aligned}
\end{equation}
The torque $\left| {{{\bm{\tau }}_G}} \right|$ is created by the $\left| {{{\bm{f}}_G}} \right|$ distributed on the circle:
\begin{equation}
\begin{aligned}
\label{equ_getorque_norm}
    \left| {{{\bm{\tau }}_G}} \right| & = \int_0^{2\pi } {D\left( \theta  \right)\frac{b}{2}\cos \theta d\theta } \\
    & = \frac{{bT}}{{2\pi }}\int_0^\pi  {\left[ {{F_G}\left( h \right) - \frac{b}{2}\sin \delta \cos \theta {\dot F_G} \left( h \right)} \right]\cos \theta d\theta } \\
    & =  - \frac{1}{8}{b^2}\sin \delta {\dot F_G} \left( h \right)T.
\end{aligned}
\end{equation}

The vector form of ${{\bm{\tau }}_G}$ in the body coordinate system can be written as
\begin{equation}
\begin{aligned}
\label{equ_getorque_vec}
    {\bf{R}}{{\bm{\tau }}_G} = \left| {{{\bm{\tau }}_G}} \right|\left( {{{\bf{z}}_B} \times {{\bf{z}}_W}} \right) = {M_G}\left( h \right)T\left( {{{\bf{z}}_B} \times {{\bf{z}}_W}} \right).
\end{aligned}
\end{equation}
${M_G}\left( h \right)$ can be regarded as

\begin{equation}
\begin{aligned}
\label{equ_Mh}
    {M_G}\left( h \right) = \frac{{{g_5}h}}{{{{\left( {{h^2} + {g_3}h + {g_4}} \right)}^2}}}.
\end{aligned}
\end{equation}

\subsubsection{Calibrate parameters with the real-world platform}


We need to validate the above model and calibrate parameters through real-world experiments.


As shown in Fig.~\ref{fig_model_valid}(e$\sim$g), on the quadrotor platform, we adjust the tilt angle of the ground board and the height of the quadrotor, increase throttle and maintain consistent speeds for all four rotors. Data on tilt angle, height, rotor speeds, and torques are collected.

From Fig.~\ref{fit_getorque_rpm} and Fig.\ref{getorque_combine} (more data in MAT.Part.4.2 \cite{groundeffectgithub}),
it is evident that ${{\bm{\tau }}_G}$ exhibits a strong correlation with both altitude $h$, tilt angle, and average rotor speed.

\begin{itemize}
\item ${{\bm{\tau }}_G}$ and $ T $: 
    Fig.~\ref{fit_getorque_rpm} illustrates the relationship between leveling torque ${{\bm{\tau }}_G}$ and average rotational speed $\left\| {{n_i}} \right\|$ under varying heights on the quadrotor platform.
    It is evident that, there is a proportional relationship between the leveling torque and the square of the average rotation speed (or thrust according to (\ref{equ_rpm_to_thrust})):

    \begin{equation}
    \begin{aligned}
    \label{equ_torque_rpm}
        \left| {{{\bm{\tau }}_G}} \right| \sim {\left\| {{n_i}} \right\|^2} \sim T.
    \end{aligned}
    \end{equation}

\item ${{\bm{\tau }}_G}$ and $ \delta  $: 
    As shown in Fig.~\ref{getorque_combine}b, the leveling torque is proportional to the tilt angle:
    \begin{equation}
    \begin{aligned}
    \label{equ_torque_angle}
        \left| {{{\bm{\tau }}_G}} \right|\sim \delta \left( {\delta  \to 0} \right).
    \end{aligned}
    \end{equation}

    However, as the tilt angle becomes larger ($\delta  > 10{\rm{^\circ }}$), the leveling torque ceases to increase.

\item ${{\bm{\tau }}_G}$ and $ h $: 
    From Fig.~\ref{getorque_combine}(a), it can be observed that, under a fixed tilt angle, as the quadrotor's altitude gradually decreases, the leveling torque initially increases and then decreases.
    This validates our previous assumption in Sec.~\ref{subsubsec:assumption} and also calibrates the parameters of (\ref{equ_Mh}).

\end{itemize}


\noindent
\textbf{\emph{Conclusion 1:}}

\textbf{As the height decreases, the absolute value of the leveling torque $\left| {{{\bm{\tau }}_G}} \right|$ reaches a maximum at a certain distance from the ground, then gradually decreases, rather than monotonically increasing.}

\begin{figure}[ht]
    \vspace{0cm}
    \begin{center}
        \includegraphics[angle=0,width=0.4\textwidth]{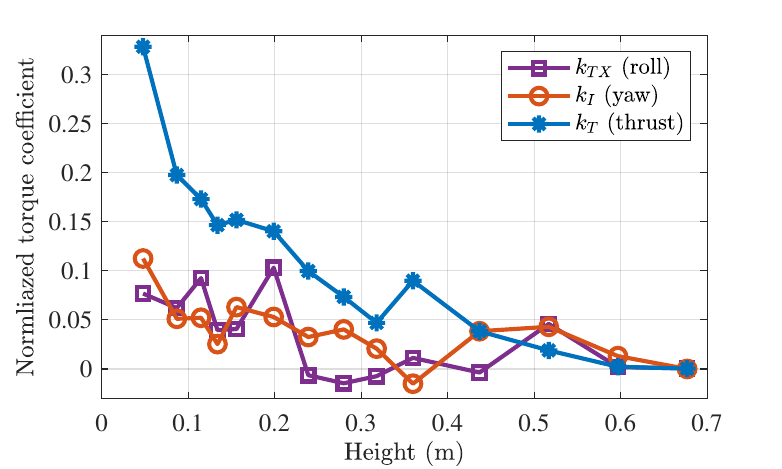}
        \caption{The relationship between normalized torque coefficient and height.}
        \label{fit_bodytorque}
    \end{center}
    \vspace{-0.8cm}
\end{figure}

\subsection{Mixing Matrix Model under Ground Effect}
\label{subsec:bodytorque}

Ground effect has various impacts on the dynamic model of the quadrotor.
We have reasons to suspect that the torque generated by rotor speed differences is also affected by ground effect, which means some parameters of the mixing control matrix in (\ref{equ_mixmatrix}) (${k_{TX}},{k_{TY}},{k_I}$) may vary with the state of the quadrotor.
Therefore, we conduct experiments to validate this.

We fix the quadrotor onto the platform (Fig.~\ref{fig_model_valid}(b$\sim$d)) and adjust the distance between the quadrotor and the ground board while ensuring the ground board is level.
Then we drive the four rotors to rotate at varying speeds.
Data on rotor speeds, heights and torques are collected.

The ground board is adjustable relative to the quadrotor's Y-axis, with a certain adjustment error that prevents the ground board from being perfectly level.
The leveling torque still exists on the Y-axis.
Therefore, when studying the torque generated by thrust difference, we only focus on the X-axis and Z-axis (${k_{TX}}, {k_I}$).



We calculate the Spearman correlation coefficient in MAT.Part.4.2 \cite{groundeffectgithub}:
${r_s}\left[ {{{\bm{x}}_W}^ \top {{\bm{\tau }}_B},{\bm{N}_{base}}\left( 2 \right)} \right] = 0.8400$, ${r_s}\left[ {{{\bm{x}}_W}^ \top {{\bm{\tau }}_B},h} \right] =  - 0.0937$.
It can be seen that the body torque ${{\bm{\tau }}_B}$ is closely related to the rotor speed difference ${\bm{N}_{base}}$ and has no relationship with the altitude $h$.



We calculate the relationship between each coefficient (${k_{T}}$, ${k_{TX}}$ and ${k_I}$) and the altitude respectively in Fig.~\ref{fit_bodytorque}.
The coefficients in the figure are normalized with the following procedure:

\begin{equation}
\begin{aligned}
\label{equ_knorm}
    \bar k\left( h \right) = \frac{{k\left( h \right)}}{{k\left( { + \infty } \right)}} - 1.
\end{aligned}
\end{equation}


In Fig.~\ref{fit_bodytorque}, as the quadrotor descends from high altitude to the ground board, the change in torque coefficient ($k_{TX}$ and $k_{I}$) generated by the rotor speed difference is negligible (less than 10\%), compared to the change in thrust coefficient $k_T$ (around 30\%).


We believe this change is mainly caused by the mechanical vibrations near the ground, leading to measurement errors.

\noindent
\textbf{\emph{Conclusion 2:}}

\textbf
{
    The torque parameters of the mixing control matrix ($k_{TX} , k_{TX} , k_{I}$) remains \emph{unaffected} by ground effect.
}

\begin{figure}[h]
    \vspace{-.02cm}
    \begin{center}
        \includegraphics[angle=0,width=0.45\textwidth]{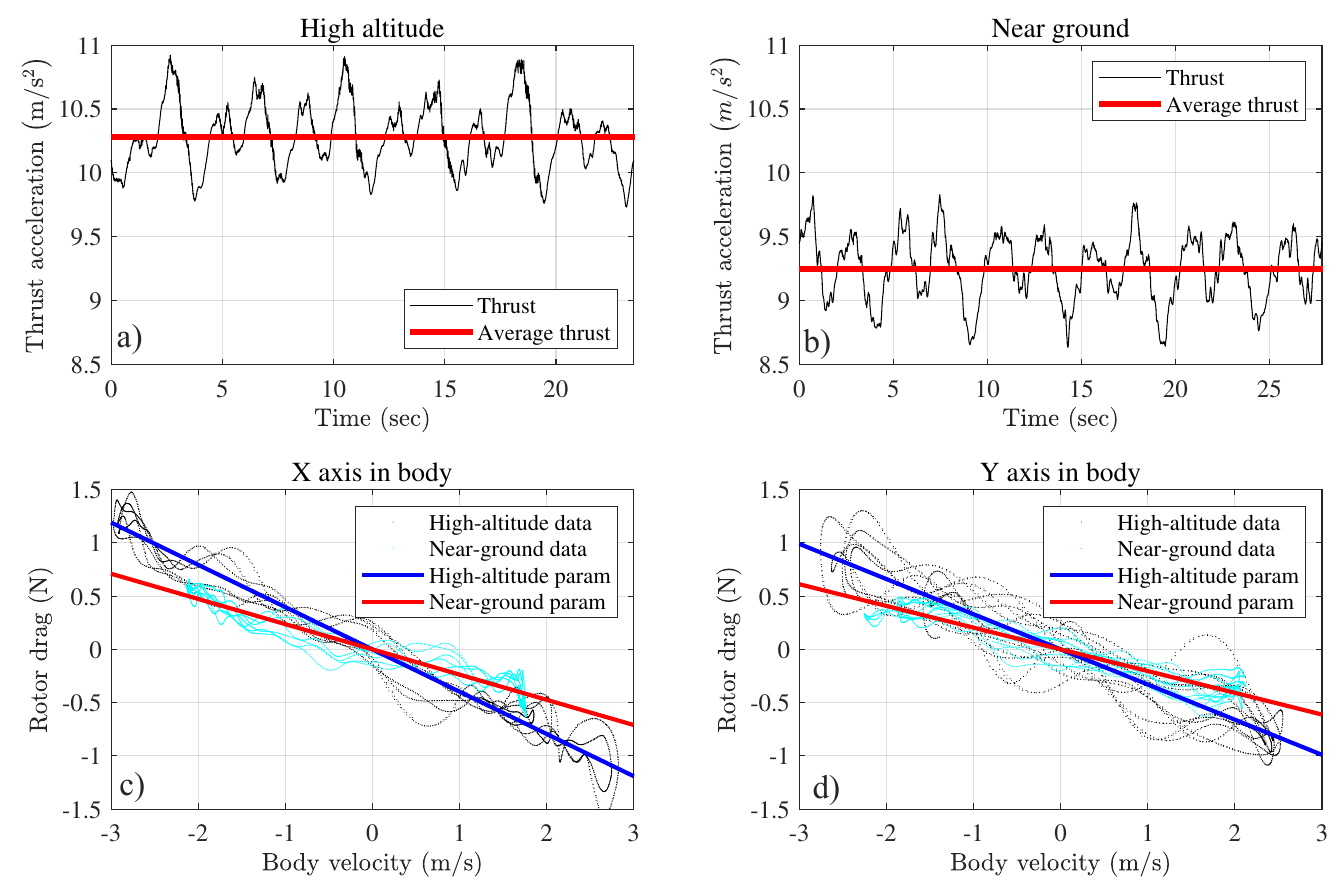}
        \caption
        {
            Thrust differences (a,b) and rotor drag (c,d) during high altitude and near ground flight.
        }
        \label{traj_drag_Tacc_comp}
    \end{center}
    \vspace{-1.0cm}
\end{figure}

\subsection{Rotor Drag under Ground Effect}
\label{subsec:geRD}


When the multicopter is flying forward, a rotor drag opposite to the direction of motion is generated.
It is proved that the drag is related to thrust and the flight speed along the horizontal direction of the multicopter's body system\cite{2017Nonlinear}:
\begin{equation}
\begin{aligned}
\label{equ_drag_with_thrust}
    {{\bm{f}}_{\bm{D}}}\left( {{\bm{R}},{\bm{v}},T} \right) =  - {\bm{RD'}}\sqrt T {{\bm{R}}^ \top }{\bm{v}},
\end{aligned}
\end{equation}
where ${\bm{D'}} = diag\left( {{d_x},{d_y},0} \right)$ is the drag coefficient matrix.


Since the thrust $T$ is in this rotor drag term, the formulation of the rotor drag poses challenges for the derivation of the differential flatness of the multicopter dynamics.
Generally, when the multicopter is flying at a high altitude, the thrust is maintained near the hover throttle, which can be regarded as a fixed value\cite{faessler2017differential}.
When the quadrotor is flying near the ground, the hover thrust is reduced due to the extra thrust provided by $\bm{f}_G$, so the rotor drag will also decrease, in theory.


To verify the assumption, we conduct flight tests at both high and low altitudes.
In this experiment (detailed data in the appendix), the ratio of the average thrust at low and high altitude is (Fig.~\ref{traj_drag_Tacc_comp}(a,b)):
    $\frac{{\sqrt {\bar T\left( {0.1} \right)} }}{{\sqrt {\bar T\left( {2.0} \right)} }} = {\rm{0}}{\rm{.9486}}$,
which should also be the ratio of rotor drag at low and high altitudes according to (\ref{equ_drag_with_thrust}). However, the experimental results in Fig.~\ref{traj_drag_Tacc_comp}(c,d) show that:
$\frac{{{d_x}\left( {0.1} \right)}}{{{d_x}\left( {2.0} \right)}} = 0.5963,\frac{{{d_y}\left( {0.1} \right)}}{{{d_y}\left( {2.0} \right)}} = 0.6179$.
So we come to a conclusion:

\noindent
\textbf{\emph{Conclusion 3:}}


\textbf{The rotor drag ${{\bf{f}}_D}$ at low altitudes declines substantially beyond what current model(\ref{equ_drag_with_thrust}) can predict.}

The formation mechanism of this phenomenon cannot be explained theoretically at the moment.
We chose to calibrate this parameters when near-ground flight is needed.

But in engineering, we can consider rotor drag as a function related to altitude and the specific curve can be measured experimentally and interpolated:

\begin{equation}
\begin{aligned}
\label{equ_ge_drag_h}
    {{\bm{f}}_D}\left( {{\bm{R}},{\bm{v}},h} \right) =  - {\bm{RD}}\left( h \right){{\bm{R}}^ \top }{\bm{v}}.
\end{aligned}
\end{equation}

\section{Differential Flatness with Ground Effect}
\label{sec:flatness}

In this section, we will explain that, even when considering the impact of ground effect, the multicopter retains its differential flatness. By designating the multicopter's position and yaw angle as its flat outputs, we can compute the multicopter's thrust, attitude, angular velocity, angular acceleration, torque, and rotor speeds:

\begin{equation}
\begin{aligned}
\label{equ_flatness}
    & \left\{ {{\bm{p}},{\bm{v}},{\bm{a}},{\bm{j}},{\bm{s}}} \right\},\left\{ {\varphi ,\dot \varphi ,\ddot \varphi } \right\}\\
    \leftrightarrow & {T_{ref}},\left\{ {{{\bm{\xi }}_{ref}},{{\bm{\omega }}_{ref}},{{{\bm{\dot \omega }}}_{ref}},{{\bm{\tau }}_{Bref}},{{\bm{n}}_{ref}}} \right\}.
\end{aligned}
\end{equation}


The more types of disturbances we need to consider, the more complex the derivation process of differential flatness becomes.
For example, when we only consider the rotor drag related to velocity\cite{faessler2017differential}, the multicopter needs to compensate for it by generating additional acceleration.
However, the acceleration of a multicopter is coupled with its attitude, which means that the compensation for drag needs to be considered from acceleration to attitude and angular velocity reference outputs.
For multicopters affected by ground effects, the derivation process needs to consider models for additional thrust, altitude-varying rotor drag, and leveling torque.


Among these disturbances, the leveling torque, which is influenced by multiple factors, is the most complex.
Although we calibrate the parameters in Sec.~\ref{subsec:getorque}, it only partially explains the model and is not entirely accurate.
There are bound to be some differences between the situation with the torque measurement platform and real flight conditions.
Therefore, we need to find a reasonable method to simplify or approximate the leveling torque model, ensuring it does not introduce distortion to the model while also not affecting the differential flatness of the multicopter.

\subsection{The Payload Model Designed for Leveling Torque}
\label{subsec:simball}

The approach we take is to equate the leveling torque to the gravity torque generated by a Payload ball rigidly connected under the multicopter.
We enumerate \textbf{four} dynamic models which are \textbf{equivalent} to each other  (refer to Fig.~\ref{inertia_h}) during inner loop attitude control.

\subsubsection{Payload, with external torque.}

\begin{figure}[h]
    \vspace{-0.2cm}
    \begin{center}
        \includegraphics[angle=0,width=0.45\textwidth]{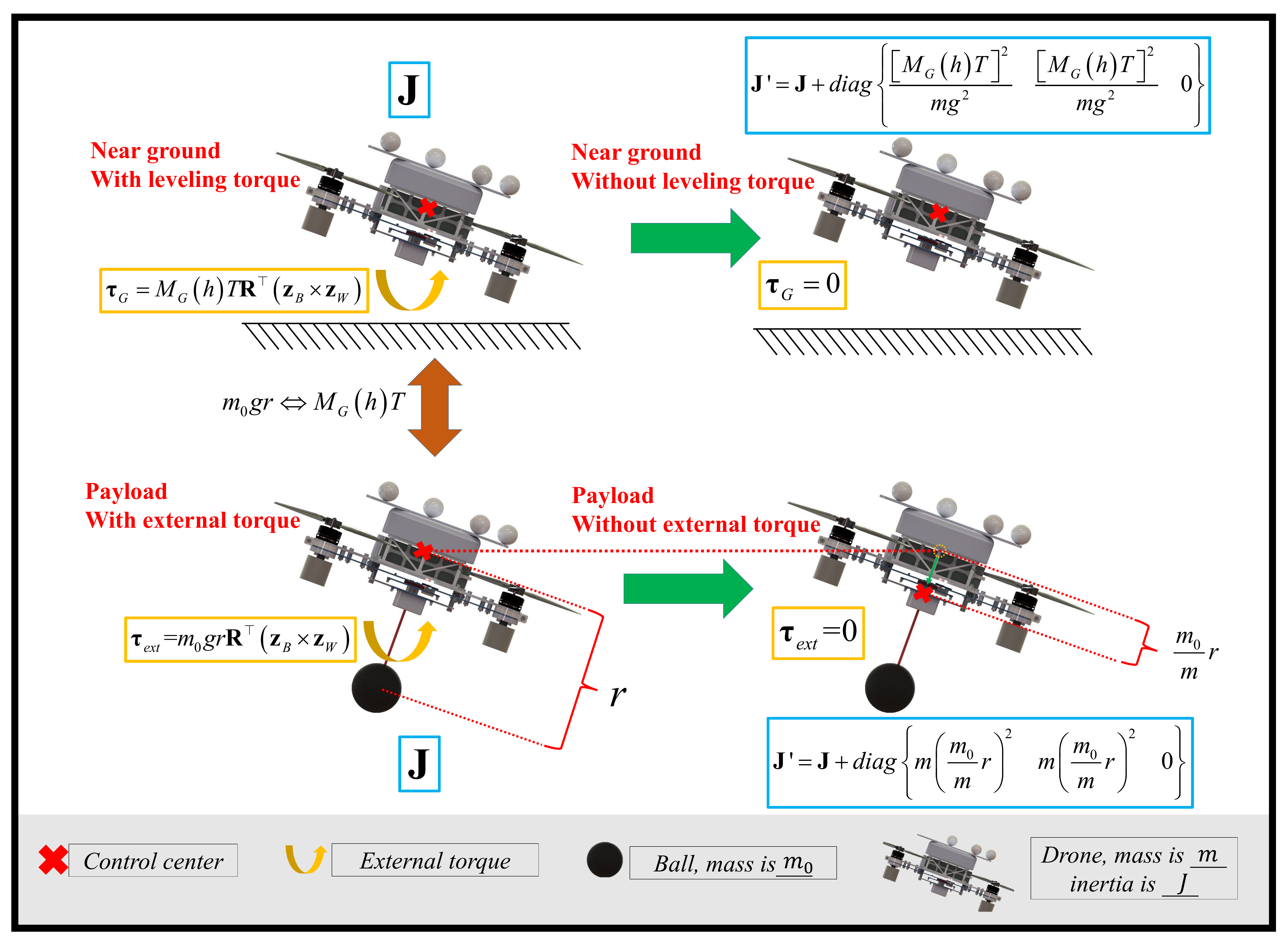}
        \caption
        {
            The payload model of leveling torque in Sec.~\ref{subsec:simball}.
        }
        \label{ballmodel}
    \end{center}
    \vspace{-0.8cm}
\end{figure}

Euler's equation for a multicopter with a payload ball (Fig.~\ref{ballmodel}) can be written as:
\begin{equation}
\begin{aligned}
\label{equ_eular_ball_simple}
    {{\bm{J}}}{\bm{\dot \omega }} =  - {\bm{\omega }} \times {{\bm{J}}}{\bm{\omega }} + {{\bm{\tau }}_B} + {m_0}gr{{\bm{R}}^ \top }\left( {{{\bm{z}}_B} \times {{\bm{z}}_W}} \right).
\end{aligned}
\end{equation}
$m$ and $m_0$ are the mass of the multicopter and the ball respectively,
$r$ is the distance from the ball to the multicopter center.

\subsubsection{Payload, without external torque.}


In the model (\ref{equ_eular_ball_simple}), if we move the control center of the multicopter from the gravity center down to the overall gravity center of the multicopter and the payload ball, the Euler equation can be written as:
\begin{equation}
\begin{aligned}
\label{equ_eular_ball_adjust}
    {{\bm{J}}'}{\bm{\dot \omega }} =  - {\bm{\omega }} \times {{\bm{J}}'}{\bm{\omega }} + {{\bm{\tau }}_B}.
\end{aligned}
\end{equation}
$J'$ is the overall inertia moment at the new control center:

\begin{equation}
\begin{aligned}
\label{equ_adjust_inertia}
    {{\bm{J}}'} = {{\bm{J}}} + diag\left\{ {\begin{array}{*{20}{c}}
        {m{{\left( {\frac{{{m_0}}}{m}r} \right)}^2}}&{m{{\left( {\frac{{{m_0}}}{m}r} \right)}^2}}&0
        \end{array}} \right\}.
\end{aligned}
\end{equation}

\subsubsection{Near ground, with leveling torque.}

Euler's equation for a multicopter with leveling torque can be written as:

\begin{equation}
\begin{aligned}
\label{equ_eular_getorque}
{\bm{J}} {\bm{\dot \omega }} =  - {\bm{\omega }} \times {\bm{J}}{\bm{\omega }} + {{\bm{\tau }}_B} + {M_G}\left( h \right)T{{\bm{R}}^ \top }\left( {{{\bm{z}}_B} \times {{\bm{z}}_W}} \right).
\end{aligned}
\end{equation}

\subsubsection{Near ground, without leveling torque.}

The form of (\ref{equ_eular_getorque}) and (\ref{equ_eular_ball_simple}) will be the same if we set

\begin{equation}
\begin{aligned}
\label{equ_eular_ball_ge}
    {m_0}gr = {M_G}\left( h \right)T.
\end{aligned}
\end{equation}

Using the same method of moving the gravity center down, we can modify the leveling torque model (\ref{equ_eular_getorque}) as:

\begin{equation}
\begin{aligned}
\label{equ_eular_ge_adjust}
    {\bm{J'\dot \omega }} =  - {\bm{\omega }} \times {\bm{J'\omega }} + {{\bm{\tau }}_B}.
\end{aligned}
\end{equation}

The adjusted inertia moment $J'$ is

\begin{equation}
\begin{aligned}
\label{equ_ge_interia_adjust}
    {\bm{J'}} = {\bm{J}} + diag\left\{ {\begin{array}{*{20}{c}}
        {\frac{{{{\left[ {{M_G}\left( h \right)T} \right]}^2}}}{{m{g^2}}}}&{\frac{{{{\left[ {{M_G}\left( h \right)T} \right]}^2}}}{{m{g^2}}}}&0
        \end{array}} \right\}.
\end{aligned}
\end{equation}

Considering that the thrust is maintained around the hover throttle, it can be assumed:

\begin{equation}
\begin{aligned}
\label{equ_MGT}
    {M_G}\left( h \right)T \approx \frac{{mg{M_G}\left( h \right)}}{{1 + {F_G}\left( h \right)}}.
\end{aligned}
\end{equation}

\begin{figure}[ht]
    \vspace{0cm}
    \begin{center}
        \includegraphics[angle=0,width=0.4\textwidth]{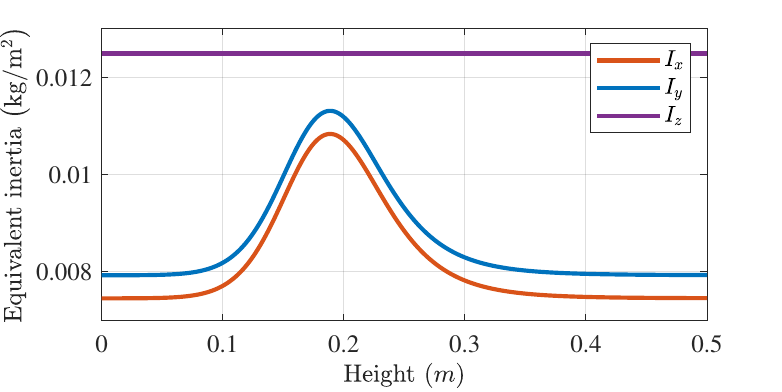}
        \caption{Equivalent inertia moment in Sec.~\ref{subsec:simball}.}
        \label{inertia_h}
    \end{center}
    \vspace{-0.8cm}
\end{figure}

Combined with (\ref{equ_ge_interia_adjust}) and (\ref{equ_MGT}), the modified inertia moment $ \bm{J}'\left( h \right) $ is only related to the height from the ground (Fig.~\ref{inertia_h}).
In order to deal with other unknown disturbances, we propose a method that combining model-based and model-free disturbance resistance, which will be introduced in detail in Sec.~\ref{sec:controller}.


\subsection{Thrust and Attitude Outputs}
\label{subsec:flat_thrust}

The model with additional thrust, altitude-varying rotor drag and inertia moment can be written as:
\begin{equation}
\begin{aligned}
\label{equ_flat_model}
    & {\bm{a}} =  - g{{\bm{z}}_W} + \left[ {1 + {F_G}\left( h \right)} \right]{T_a}{{\bm{z}}_B} - {\bm{R}}{{\bm{D}}_a}\left( h \right){{\bm{R}}^ \top }{\bm{v}}\\
    & {\bm{J}'}\left( h \right){\bm{\dot \omega }} =  - {\bm{\omega }} \times {\bm{J}'}\left( h \right){\bm{\omega }} + {{\bm{\tau }}_B},
\end{aligned}
\end{equation}
where ${T_a} = T/m$ and ${{\bm{D}}_a}\left( h \right) = {\bm{D}}\left( h \right)/m$ are the normalized accelerations.
Left multiply $\bm{z}_b$ to the following formula:
\begin{equation}
\begin{aligned}
\label{equ_flat_xyz}
    {\bm{a}} + g{{\bm{z}}_W} & + \left[ {{d_{ax}}\left( h \right){{\bm{x}}_B}^ \top {\bm{v}}} \right]{{\bm{x}}_{\bm{B}}} + \left[ {{d_{ay}}\left( h \right){{\bm{y}}_B}^ \top {\bm{v}}} \right]{{\bm{y}}_B}\\
    &  - \left[ {1 + {F_G}\left( h \right)} \right]{T_a}{{\bm{z}}_B} = 0.
\end{aligned}
\end{equation}
The reference thrust can be obtained:
\begin{equation}
\begin{aligned}
\label{equ_flat_thrust}
    T = m\frac{{{{\bm{z}}_B}^ \top \left( {{\bm{a}} + g{{\bm{z}}_W}} \right)}}{{\left[ {1 + {F_G}\left( h \right)} \right]}}.
\end{aligned}
\end{equation}

The derivation of the reference attitude can refer to the previous works \cite{faessler2017differential,wang2022geometrically}.





\subsection{Other Outputs}
\label{subsec:flat_other}


We can obtain the reference outputs of angular velocity $ {{\bm{\omega }}_{ref}} $ and angular acceleration $ {{{\bm{\dot \omega }}}_{ref}} $ by differentiating the acceleration in (\ref{equ_flat_model})
The detailed derivation process is similar to the previous works \cite{faessler2017differential, wang2022geometrically}.
In the differentiation process, since the multicopter generally does not have a significant Z-axis velocity when executing a near-ground flight trajectory, we can assume that
$ \frac{d}{{dt}}D\left( h \right) = \frac{{{d^2}}}{{d{t^2}}}D\left( h \right) = 0 $.

Then the reference torque input can be obtained:

\begin{equation}
\begin{aligned}
\label{equ_flat_torque}
    {{\bm{\tau }}_{ref}} = {\bm{J}'}\left( h \right){{{\bm{\dot \omega }}}_{ref}} + {{\bm{\omega }}_{ref}} \times {\bm{J}'}\left( h \right){{\bm{\omega }}_{ref}}.
\end{aligned}
\end{equation}

${\bm{J}'}\left( h \right)$ uses the conclusion from (\ref{equ_ge_interia_adjust})(\ref{equ_MGT}), illustrating the effect of leveling torque model on the differential flatness.

\section{Model-based Controller with Incremental Inversion}
\label{sec:controller}


This section will use the previously introduced model to compensate for external disturbances, thereby controlling the quadrotor during near-ground flight.
The most important part is the compensation of the leveling torque in Sec.~\ref{subsec:thrust_torque}.

\subsection{Acceleration Command}
\label{subsec:acccmd}



Given that we establish an accurate external force model, we rely on these models to compensate disturbances to reduce control errors.
The expected resultant acceleration is
\begin{equation}
\begin{aligned}
\label{equ_total_acc}
    {{\bm{a}}_{des}} & = {{\bm{a}}_{ref}} + {{\bm{a}}_E} - {{\bm{a}}_D} - {{\bm{a}}_G}, \\
    {{\bm{a}}_E} & = {{\bm{K}}^P}\left( {{{\bm{p}}_{des}} - {\bm{\hat p}}} \right) + {{\bm{K}}^V}\left( {{{\bm{v}}_{des}} - {\bm{\hat v}}} \right), \\
    {{\bm{a}}_D} & =  - {\bm{R}}\frac{{{\bm{D}}\left( h \right)}}{m}{{\bm{R}}^ \top }{{\bm{v}}_{ref}}, \\
    {{\bm{a}}_G} & = \frac{{{T_{ref}}}}{m}{F_G}\left( h \right){{\bm{z}}_B}.    
\end{aligned}
\end{equation}
${{\bm{a}}_{ref}} $ represents the reference acceleration (\ref{equ_flatness}) along the motion trajectory, 
${{\bm{a}}_E}$ is the acceleration resulting from the position and velocity error feedback, 
(${\bm{K}}^P$ and ${\bm{K}}^V$ are the parameters of the PD controller for position control.),
${{\bm{a}}_D}$ and ${{\bm{a}}_G}$ are the accelerations from the drag (\ref{equ_ge_drag_h}) and additional thrust model(\ref{equ_geforce}) respectively.

\subsection{Attitude, Bodyrate and Body Acceleration Command}

We can get the desired attitude from the desired acceleration using the principles of differential flatness in Sec.~\ref{sec:flatness}:

\begin{equation}
\begin{aligned}
\label{equ_q_cmd}
    {{\bm{a}}_{des}} \leftrightarrow {{\bm{\xi }}_{des}}.
\end{aligned}
\end{equation}

The error between the current attitude ${{\bm{\hat \xi }}}$ and the expected attitude ${{\bm{\xi }}_{des}}$ is:

\begin{equation}
\begin{aligned}
\label{equ_quat_qerror}
    {{\bm{\xi }}_e} = {\bm{\hat \xi }} \circ {{\bm{\xi }}_{des}}.
\end{aligned}
\end{equation}
The attitudes here are all quaternion forms. The $ \circ $ means the Hamilton quaternion product.
Write it into an angle vector:
\begin{equation}
\begin{aligned}
\label{equ_angle_qerror}
    {{\bm{\xi }}_e}' = \frac{{2{{\cos }^{ - 1}}{\bm{\xi }}_{e}^w}}{{\sqrt {1 - {\bm{\xi }}{{_{e}^w}^2}} }}{\left[ {\begin{array}{*{20}{c}}
    {{\bm{\xi }}_{e}^x}&{{\bm{\xi }}_{e}^y}&{{\bm{\xi }}_{e}^z}
    \end{array}} \right]^ \top },
\end{aligned}
\end{equation}
where $\left[ {\begin{array}{*{20}{c}}
    {{{\bm{\xi }}^w}}&{{{\bm{\xi }}^x}}&{{{\bm{\xi }}^y}}&{{{\bm{\xi }}^z}}
    \end{array}} \right]$
is the quaternion form of the attitude.
Subsequently, we can calculate the desired angular velocity and angular acceleration:
\begin{equation}
\begin{aligned}
\label{equ_bodyrate_cmd}
    {{\bm{\omega }}_{des}} & = {{\bm{K}}^{\bm{\xi }}}{{\bm{\xi }}_e}' + {{\bm{\omega }}_{ref}} \\
    {{{\bm{\dot \omega }}}_{des}} & = {{\bm{K}}^\omega }\left( {{{\bm{\omega }}_{des}} - {{\bm{\omega }}_f}} \right) + {{{\bm{\dot \omega }}}_{ref}}.
\end{aligned}
\end{equation}
In this context, $ {{\bm{K}}^{\bm{\xi }}} $ and $ {{\bm{K}}^\omega } $ are the PD parameters for angle control, and $ {{\bm{\omega }}_{ref}} $, $ {{{\bm{\dot \omega }}}_{ref}} $  are the reference angular velocity and angular acceleration derived from (\ref{equ_flatness}) respectively.

\subsection{Thrust and Torque Command}
\label{subsec:thrust_torque}
The desired acceleration projected onto the quadrotor's Z-axis corresponds to the acceleration the rotors need to produce. Then we can get the expected thrust:

\begin{equation}
\begin{aligned}
\label{equ_acc_cmd}
    {T_{des}} = m{{\bm{a}}_{des}}\frac{{{{{\bm{\hat z}}}_B}}}{{\left| {{{{\bm{\hat z}}}_B}} \right|}}.
\end{aligned}
\end{equation}

The torque required from the motors can be calculated based on the desired angular velocity and desired angular acceleration:
\begin{equation}
\begin{aligned}
\label{equ_torque_cmd}
    {}^B{{\bm{\tau }}_{des}} = {\bm{J}'}\left( h \right) \cdot {{{\bm{\dot \omega }}}_{des}} + {{\bm{\omega }}_{des}} \times {\bm{J}'}\left( h \right) \cdot {{\bm{\omega }}_{des}}.
\end{aligned}
\end{equation}
The rotational inertia $ {\bm{J}'}\left( h \right) $, which is altitude-dependent, serves to compensate for the torque effects induced by ground effect.


Due to various reasons, the above method cannot completely compensate for the external torques.
Firstly, the parameters obtained based on the torque measurement platform are not entirely accurate.
Secondly, there are other unmodeled torques, such as the gravity torque and the gyroscopic torque.
Therefore, we combine the incremental nonlinear dynamic inversion \cite{tal2020accurate} with our model to output the control torque:

\begin{equation}
\begin{aligned}
\label{equ_torque_cmd_indi}
    {}^B{{\bm{\tau }}_{des}} = {{{\bm{\hat \tau }}}_B} + {\bm{J}'}\left( h \right) \cdot \left( {{{{\bm{\dot \omega }}}_{des}} - {{{\bm{\dot \omega }}}_f}} \right).
\end{aligned}
\end{equation}

In practice, we still need to fine-tune the function $ {\bm{J}'}\left( h \right) $ during the experiment.
The model-based compensation method then degrades to adjust the gain of the angular acceleration error.
This approach is equivalent to setting the angular velocity control gain in a conventional dual-loop attitude controller as a height-dependent function. It is also similar to a \textit{piecewise PID controller}, making it highly feasible for engineering applications.

\subsection{Rotor Speed Command}

Based on the inverse of the mixed control matrix in (\ref{equ_mixmatrix}), the rotational speeds of each motor can be computed from the thrust and torques:

\begin{equation}
\begin{aligned}
\label{equ_rpm_cmd}
    {{\bm{N}}_{des}}^2 = {M^{ - 1}}\left[ {\begin{array}{*{20}{c}}
    {{T_{des}}}\\
    {{}^B{{\bm{\tau }}_{des}}}
    \end{array}} \right].
\end{aligned}
\end{equation}


The rotor speeds are controlled through the throttle input ($t_i^{des} \in \left[ {0,1} \right]$), and the rotational speeds are fed back through \emph{BDhot}.
Methods of motor modeling \cite{brandt2011propeller,zhang2022visual}, calibration and speed control are described in MAT.Part.3 \cite{groundeffectgithub}.

We refine the code of the APM firmware \cite{groundeffectgithub}
to align with the throttle interface and voltage compensation logic, which has been open-sourced \cite{groundeffectgithub}.

\begin{table*}[h]
\centering
\caption{Comparison of controllers.}
\label{tab:ctrl}
\begin{tabular}{ccccc|ccccc}
\toprule[2pt]
\multirow{2}{*}{Experiment}
& Acceleration
& Torque
& \multirow{2}{*}{Height}
& \multirow{2}{*}{Velocity}
& \multicolumn{3}{c}{RMSE (cm)}
& $ \max \left( {{{\left\| {{{\bm{E}}_P}} \right\|}_2}} \right) $
& $ \sigma \left( {{{\left\| {{{\bm{E}}_P}} \right\|}_2}} \right) $                                                                                         \\
  & Compensation        & Compensation      &           &           &XOY            & Z             & All           & (cm)              & (cm)              \\ \midrule[2pt]
1 & -                   & -                 & 0.12m     & 1m/s      & 6.54          & 8.64          & 10.85         & 18.04             & 2.19              \\
2 & INDI                & -                 & 0.12m     & 1m/s      & \textbf{5.51} & 6.83          & 8.77          & \textbf{16.07}    & \textbf{2.06}     \\
3 & Neural Land         & -                 & 0.12m     & 1m/s      & 6.81          & 4.94          & 8.41          & 16.75             & 2.73              \\
4 & Proposed            & -                 & 0.12m     & 1m/s      & 7.12          & \textbf{1.94} & \textbf{7.38} & 16.45             & 3.05              \\
5 & Proposed            & Proposed          & 0.12m     & 1m/s      & \textbf{4.35} & \textbf{0.97} & \textbf{4.45} & \textbf{11.97}    & 2.32              \\ \midrule[1pt]
6 & Proposed            & -                 & 0.18m     & 3m/s      & 12.63         & \textbf{1.71} & 12.74         & 26.27             & 5.36              \\
7 & Proposed            & Proposed          & 0.18m     & 3m/s      & \textbf{6.67} & 2.84          & \textbf{6.97} & \textbf{14.66}    & \textbf{2.72}     \\ \midrule[1pt]
8 & Proposed            & Proposed          & 1.75m     & 5m/s      & \textbf{4.77} & \textbf{1.41} & \textbf{4.98} & \textbf{9.75}     & \textbf{1.64}     \\ \bottomrule[2pt]
\end{tabular}
\vspace{-0.5cm}
\end{table*}

\begin{figure}[ht]
    \vspace{-0.5cm}
    \begin{center}
        \includegraphics[angle=0,width=0.45\textwidth]{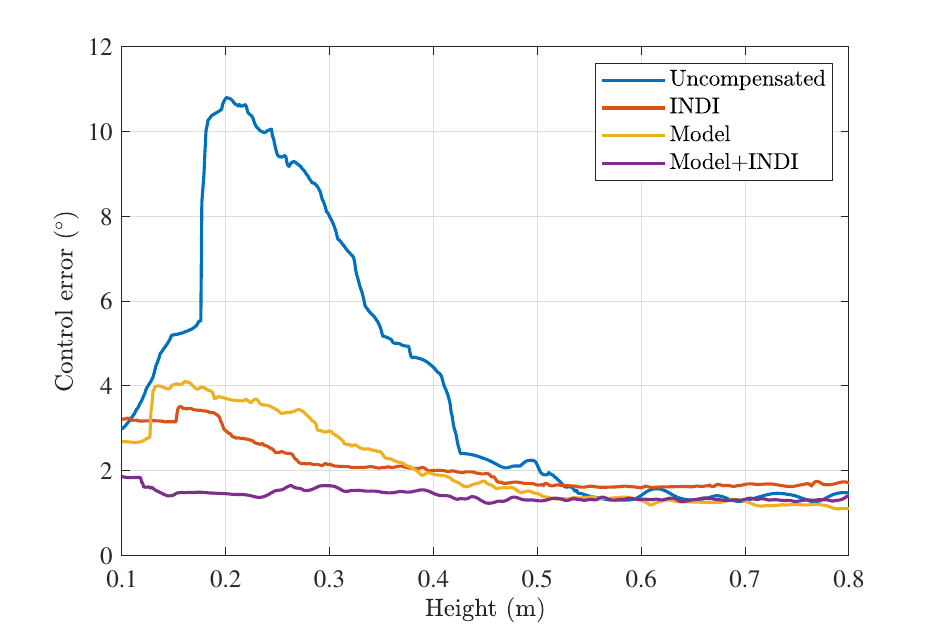}
        \caption{Hovering angle error with different control methods.}
        \label{traj_angle_rmse}
    \end{center}
    \vspace{-1cm}
\end{figure}

\section{Flight Experiment}
\label{sec:experiment}


We set up a quadrotor flight platform for comparison experiments.
The quadrotor has a larger blade diameter ($r = \rm{7{\rm{inch}}}$) and a smaller propeller distance to the ground (${h_{\min }} = \rm{7.2cm}$). These factors make the ground effect of the quadrotor more obvious.
The controller in Sec.~\ref{sec:controller} is compared with other control methods through hovering and trajectory tracking experiments.


The quadrotor is equipped with a flight controller (CUAV V5+\cite{groundeffectgithub})
that can feedback $\rm{200Hz}$ IMU data.
The flight controller using an APM firmware version that supports \emph{BDhot}
can communicate bidirectionally with the motor controller and read the rotor speed data at $\rm{200Hz}$.
A $16 \times 16$ lattice laser sensor \cite{groundeffectgithub}
is mounted under the quadrotor to measure the distance to the ground (at $\rm{100Hz}$).
An Intel NUC is used as the computing platform.
A label holder for the NOKOV Motion Capture System \cite{groundeffectgithub} is fixed to the top of the computer.
The motion capture system and IMU are fused through Extended Kalman Filter (EKF)\cite{madyastha2011extended, sola2017quaternion}, and a smooth location result ($\rm{200Hz}$) is obtained.

\subsection{Hovering Experiment}
\label{subsec:ctrl}


We hover the quadrotor from high to low altitude and compare the angle control errors.
The error value at a specific altitude is defined as:

\begin{equation}
\begin{aligned}
\label{equ_angleerror}
    E\left( {{h_0}} \right) = {\left\| {{\bm{\hat \xi }} - {{\bm{\xi }}_{des}}} \right\|_2}\left( {h \in \left[ {{h_0} - \Delta h,{h_0} + \Delta h} \right]} \right).
\end{aligned}
\end{equation}


Fig.~\ref{traj_angle_rmse} shows how the control error varies with altitude.
When no compensation is applied, the angle error at low altitude is huge, reaching its peak at $h = 0.2m$.
This corresponds to the model in Sec.~\ref{subsec:getorque}, where the leveling torque in Fig.\ref{getorque_combine} reaches its maximum at $h = 0.18m$.


Both the model-based method in (\ref{equ_torque_cmd}) and INDI \cite{tal2020accurate} can effectively reduce the control error, but the angle error near the ground is still increased compared with the high altitude. When the combined model and incremental inverse method (\ref{equ_torque_cmd_indi}) is used, the error at low altitude is the same as that at high altitude.

\subsection{Trajectory Tracking Experiment}

We generate lemniscate trajectories \cite{wang2022geometrically} for tracking experiments in TABLE.~\ref{tab:ctrl}.


Executing the trajectory at $v=1m/s$ without torque compensation, we conduct four sets of experiments to compare our acceleration compensation method with INDI \cite{tal2020accurate}, neural landing\cite{shi2019neural}, and the no-compensation method\cite{faessler2017differential}.


INDI performs only slightly better than no compensation in the Z-axis.
This could be because INDI compensates for the acceleration of external forces with real-time estimation.
However, a slight error in compensation for the external force in the Z-axis can affect the multicopter's altitude, and the altitude change will further affect the external force in the Z-axis, which leads to a static control error.


The reasonable approach is to compensate with the predicted external force under the desired state, as seen in the proposed method.
The proposed method performs the best in the Z-axis, which is expected, as the feedforward curve of ${{\bm{f}}_G}$ is finely calibrated based on the multicopter.
In these four sets of experiments, INDI has the best control performance in the XOY plane.
This might be because the attitude cannot track the acceleration perfectly due to the external torques, but INDI can introduce acceleration into the feedback loop for compensation.


With the torque compensation, the proposed method achieves the best control accuracy at $1m/s$ and $3m/s$ trajectories.
To assess the controller's extreme performance, we conduct a trajectory test at a high altitude of $\rm{5m/s}$. Our controller achieves an RMSE of $\rm{4.98}cm$. The controller does perform well, but low-altitude flight still hurts trajectory tracking.
Please check MAT.Part.5 \cite{groundeffectgithub} for detailed experimental data.

\section{Discussion and future work}
\label{sec:conclusion}


This paper summarizes various models of the multicopter under ground effect and establishes a control methodology. We consider the additional thrust, leveling torque, and rotor drag models under ground effect and validate its impact on the mixing matrix. However, our study has several limitations.
Firstly, we do not model the thrust decrease during high-speed forward flight\cite{kan2019analysis}. This is mainly due to two reasons: our models are primarily calibrated at static force platforms, and our flight speed does not reach the threshold that would produce a significant loss of lift.
In our flight scenarios, the flight speed remains below the induced speed of $4m/s$, and the minimum flight altitude is larger than the propeller radius, resulting in the lift reduction being not obvious and, therefore, not considered.
However, Exp.6 and Exp.7 in TABLE.~\ref{tab:ctrl} indicate that high-speed flight impacts the control effectiveness along the Z-axis.
Secondly, our study on the decrease of rotor drag is not thorough enough. The causes of rotor drag involve complex airflow phenomena \cite{bristeau2009role}, and the ground effect further complicates the situation. We hope that future research will explain this phenomenon clearly.


\bibliography{references}

\begin{thebibliography}{10}
\providecommand{\url}[1]{#1}
\csname url@samestyle\endcsname
\providecommand{\newblock}{\relax}
\providecommand{\bibinfo}[2]{#2}
\providecommand{\BIBentrySTDinterwordspacing}{\spaceskip=0pt\relax}
\providecommand{\BIBentryALTinterwordstretchfactor}{4}
\providecommand{\BIBentryALTinterwordspacing}{\spaceskip=\fontdimen2\font plus
\BIBentryALTinterwordstretchfactor\fontdimen3\font minus
  \fontdimen4\font\relax}
\providecommand{\BIBforeignlanguage}[2]{{%
\expandafter\ifx\csname l@#1\endcsname\relax
\typeout{** WARNING: IEEEtran.bst: No hyphenation pattern has been}%
\typeout{** loaded for the language `#1'. Using the pattern for}%
\typeout{** the default language instead.}%
\else
\language=\csname l@#1\endcsname
\fi
#2}}
\providecommand{\BIBdecl}{\relax}
\BIBdecl

\bibitem{fishman2021dynamic}
J.~Fishman, S.~Ubellacker, N.~Hughes, and L.~Carlone, ``Dynamic grasping with
  a" soft" drone: From theory to practice,'' in \emph{2021 IEEE/RSJ
  International Conference on Intelligent Robots and Systems (IROS)}, 2021, pp.
  4214--4221.

\bibitem{saunders2024autonomous}
J.~Saunders, S.~Saeedi, and W.~Li, ``Autonomous aerial robotics for package
  delivery: A technical review,'' \emph{Journal of Field Robotics}, vol.~41,
  no.~1, pp. 3--49, 2024.

\bibitem{gao2019exploiting}
S.~Gao, C.~Di~Franco, D.~Carter, D.~Quinn, and N.~Bezzo, ``Exploiting ground
  and ceiling effects on autonomous uav motion planning,'' in \emph{2019 IEEE
  International Conference on Unmanned Aircraft Systems (ICUAS)}, 2019, pp.
  768--777.

\bibitem{wang2022neither}
L.~Wang, H.~Xu, Y.~Zhang, and S.~Shen, ``Neither fast nor slow: How to fly
  through narrow tunnels,'' \emph{IEEE Robotics and Automation Letters},
  vol.~7, no.~2, pp. 5489--5496, 2022.

\bibitem{ding2022passive}
R.~Ding, Y.-H. Hsiao, H.~Jia, S.~Bai, and P.~Chirarattananon, ``Passive wall
  tracking for a rotorcraft with tilted and ducted propellers using proximity
  effects,'' \emph{IEEE Robotics and Automation Letters}, vol.~7, no.~2, pp.
  1581--1588, 2022.

\bibitem{luo2015biomimetic}
C.~Luo, X.~Li, Y.~Li, and Q.~Dai, ``Biomimetic design for unmanned aerial
  vehicle safe landing in hazardous terrain,'' \emph{IEEE/ASME Transactions on
  Mechatronics}, vol.~21, no.~1, pp. 531--541, 2015.

\bibitem{ji2022real}
J.~Ji, T.~Yang, C.~Xu, and F.~Gao, ``Real-time trajectory planning for aerial
  perching,'' in \emph{2022 IEEE/RSJ International Conference on Intelligent
  Robots and Systems (IROS)}, 2022, pp. 10\,516--10\,522.

\bibitem{powers2013influence}
C.~Powers, D.~Mellinger, A.~Kushleyev, B.~Kothmann, and V.~Kumar, ``Influence
  of aerodynamics and proximity effects in quadrotor flight,'' in
  \emph{Experimental Robotics: The 13th International Symposium on Experimental
  Robotics}.\hskip 1em plus 0.5em minus 0.4em\relax Springer, 2013, pp.
  289--302.

\bibitem{sanchez2017characterization}
P.~Sanchez-Cuevas, G.~Heredia, and A.~Ollero, ``Characterization of the
  aerodynamic ground effect and its influence in multirotor control,''
  \emph{International Journal of Aerospace Engineering}, vol. 2017, 2017.

\bibitem{he2019ground}
X.~He, G.~Kou, M.~Calaf, and K.~K. Leang, ``In-ground-effect modeling and
  nonlinear-disturbance observer for multirotor unmanned aerial vehicle
  control,'' \emph{Journal of Dynamic Systems, Measurement, and Control}, vol.
  141, no.~7, 2019.

\bibitem{wei2019mitigating}
P.~Wei, S.~N. Chan, S.~Lee, and Z.~Kong, ``Mitigating ground effect on mini
  quadcopters with model reference adaptive control,'' \emph{International
  Journal of Intelligent Robotics and Applications}, vol.~3, no.~3, pp.
  283--297, 2019.

\bibitem{bristeau2009role}
P.-J. Bristeau, P.~Martin, E.~Sala{\"u}n, and N.~Petit, ``The role of propeller
  aerodynamics in the model of a quadrotor uav,'' in \emph{2009 IEEE European
  control conference (ECC)}, 2009, pp. 683--688.

\bibitem{faessler2017differential}
M.~Faessler, A.~Franchi, and D.~Scaramuzza, ``Differential flatness of
  quadrotor dynamics subject to rotor drag for accurate tracking of high-speed
  trajectories,'' \emph{IEEE Robotics and Automation Letters}, vol.~3, no.~2,
  pp. 620--626, 2017.

\bibitem{svacha2017improving}
J.~Svacha, K.~Mohta, and V.~Kumar, ``Improving quadrotor trajectory tracking by
  compensating for aerodynamic effects,'' in \emph{2017 international
  conference on unmanned aircraft systems (ICUAS)}.\hskip 1em plus 0.5em minus
  0.4em\relax IEEE, 2017, pp. 860--866.

\bibitem{kan2019analysis}
X.~Kan, J.~Thomas, H.~Teng, H.~G. Tanner, V.~Kumar, and K.~Karydis, ``Analysis
  of ground effect for small-scale uavs in forward flight,'' \emph{IEEE
  Robotics and Automation Letters}, vol.~4, no.~4, pp. 3860--3867, 2019.

\bibitem{cheeseman1955effect}
I.~Cheeseman and W.~Bennett, ``The effect of the ground on a helicopter rotor
  in forward flight,'' 1955.

\bibitem{TRIPATHI20173680}
\BIBentryALTinterwordspacing
A.~K. Tripathi, V.~V. Patel, and R.~Padhi, ``Autonomous landing of uavs under
  unknown disturbances using ndi autopilot with l1 adaptive augmentation,''
  \emph{IFAC-PapersOnLine}, vol.~50, no.~1, pp. 3680--3684, 2017, 20th IFAC
  World Congress. [Online]. Available:
  \url{https://www.sciencedirect.com/science/article/pii/S2405896317309321}
\BIBentrySTDinterwordspacing

\bibitem{du2016advanced}
H.~Du, Z.~Pu, J.~Yi, and H.~Qian, ``Advanced quadrotor takeoff control based on
  incremental nonlinear dynamic inversion and integral extended state
  observer,'' in \emph{2016 IEEE Chinese Guidance, Navigation and Control
  Conference (CGNCC)}, 2016, pp. 1881--1886.

\bibitem{YuCompensating}
P.~Yu, Y.~Su, L.~Ruan, and T.-C. Tsao, ``Compensating aerodynamics of
  over-actuated multi-rotor aerial platform with data-driven iterative learning
  control,'' \emph{IEEE Robotics and Automation Letters}, vol.~8, no.~10, pp.
  6187--6194, 2023.

\bibitem{SAAT202574}
\BIBentryALTinterwordspacing
S.~Saat, M.~A. Ahmad, and M.~R. Ghazali, ``Data-driven brain emotional
  learning-based intelligent controller-pid control of mimo systems based on a
  modified safe experimentation dynamics algorithm,'' \emph{International
  Journal of Cognitive Computing in Engineering}, vol.~6, pp. 74--99, 2025.
  [Online]. Available:
  \url{https://www.sciencedirect.com/science/article/pii/S2666307424000494}
\BIBentrySTDinterwordspacing

\bibitem{KAIDataDriven}
T.~KAI and R.~KAKURAI, ``A data-driven gain tuning method for automatic
  hovering control of multicopters via just-in-time modeling,'' \emph{IEICE
  Transactions on Fundamentals of Electronics, Communications and Computer
  Sciences}, vol. E106.A, 08 2022.

\bibitem{shi2019neural}
G.~Shi, X.~Shi, M.~O’Connell, R.~Yu, K.~Azizzadenesheli, A.~Anandkumar,
  Y.~Yue, and S.-J. Chung, ``Neural lander: Stable drone landing control using
  learned dynamics,'' in \emph{2019 IEEE International Conference on Robotics
  and Automation (ICRA)}, 2019, pp. 9784--9790.

\bibitem{he2017modeling}
X.~He, M.~Calaf, and K.~K. Leang, ``Modeling and adaptive nonlinear disturbance
  observer for closed-loop control of in-ground-effects on multi-rotor uavs,''
  in \emph{Dynamic Systems and Control Conference}, vol. 58295.\hskip 1em plus
  0.5em minus 0.4em\relax American Society of Mechanical Engineers, 2017, p.
  V003T39A004.

\bibitem{tal2020accurate}
E.~Tal and S.~Karaman, ``Accurate tracking of aggressive quadrotor trajectories
  using incremental nonlinear dynamic inversion and differential flatness,''
  \emph{IEEE Transactions on Control Systems Technology}, vol.~29, no.~3, pp.
  1203--1218, 2020.

\bibitem{de2012jackknife}
M.~de~Carvalho and F.~J. Marques, ``Jackknife euclidean likelihood-based
  inference for spearman's rho,'' \emph{North American Actuarial Journal},
  vol.~16, no.~4, pp. 487--492, 2012.

\bibitem{groundeffectgithub}
\BIBentryALTinterwordspacing
YangTiankai, ``Ground-effect-controller.'' [Online]. Available:
  \url{https://github.com/ZJU-FAST-Lab/Ground-effect-controller}
\BIBentrySTDinterwordspacing

\bibitem{2017Nonlinear}
J.~M. Kai, G.~Allibert, M.~Hua, and T.~Hamel, ``Nonlinear feedback control of
  quadrotors exploiting first-order drag effects,'' \emph{Ifac Papersonline},
  vol.~50, no.~1, pp. 8189--8195, 2017.

\bibitem{wang2022geometrically}
Z.~Wang, X.~Zhou, C.~Xu, and F.~Gao, ``Geometrically constrained trajectory
  optimization for multicopters,'' \emph{IEEE Transactions on Robotics},
  vol.~38, no.~5, pp. 3259--3278, 2022.

\bibitem{brandt2011propeller}
J.~Brandt and M.~Selig, ``Propeller performance data at low reynolds numbers,''
  in \emph{49th AIAA Aerospace Sciences Meeting including the New Horizons
  Forum and Aerospace Exposition}, 2011, p. 1255.

\bibitem{zhang2022visual}
K.~Zhang, T.~Yang, Z.~Ding, S.~Yang, T.~Ma, M.~Li, C.~Xu, and F.~Gao, ``The
  visual-inertial-dynamical multirotor dataset,'' in \emph{2022 IEEE
  International Conference on Robotics and Automation (ICRA)}, 2022, pp.
  7635--7641.

\bibitem{madyastha2011extended}
V.~Madyastha, V.~Ravindra, S.~Mallikarjunan, and A.~Goyal, ``Extended kalman
  filter vs. error state kalman filter for aircraft attitude estimation,'' in
  \emph{AIAA Guidance, Navigation, and Control Conference}, 2011, p. 6615.

\bibitem{sola2017quaternion}
J.~Sola, ``Quaternion kinematics for the error-state kalman filter,''
  \emph{arXiv preprint arXiv:1711.02508}, 2017.

\end{thebibliography}


\vspace{-1.8cm}
\begin{IEEEbiography}[{\includegraphics[width=1in,height=1.0in,clip,keepaspectratio]{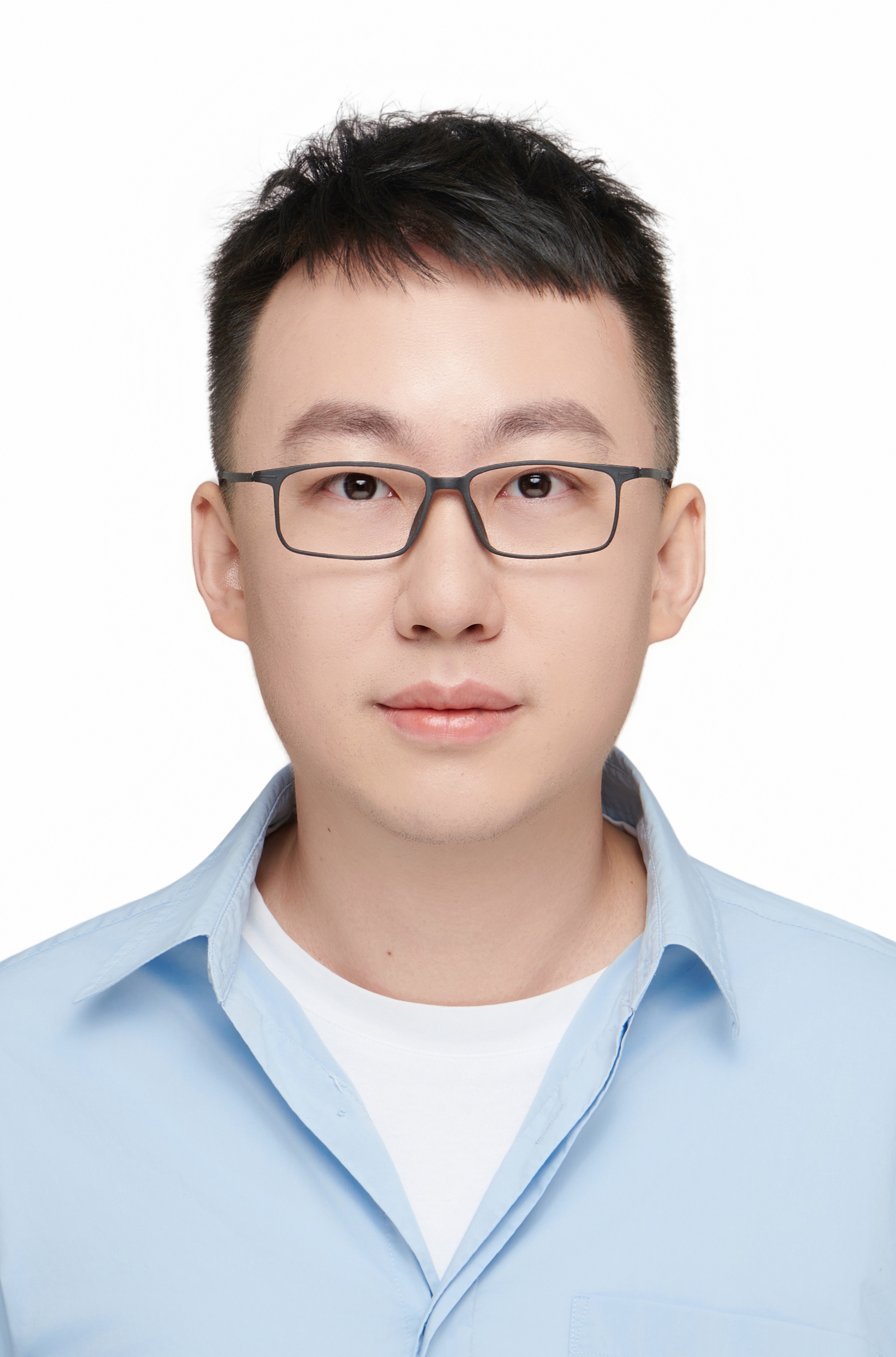}}]{Tiankai Yang}
graduated from Northwestern Polytechnical University in 2019 with a major in Detection, Guidance, and Control Technology. He is currently pursuing a Ph.D. student in Control Science and Engineering at the FAST Lab, Zhejiang University, under the supervision of Professor Fei Gao. His research focuses on control and navigation for aerial robotics and unmanned vehicles.
\end{IEEEbiography}

\vspace{-2.2cm}
\begin{IEEEbiography}[{\includegraphics[width=1in,height=1.0in,clip,keepaspectratio]{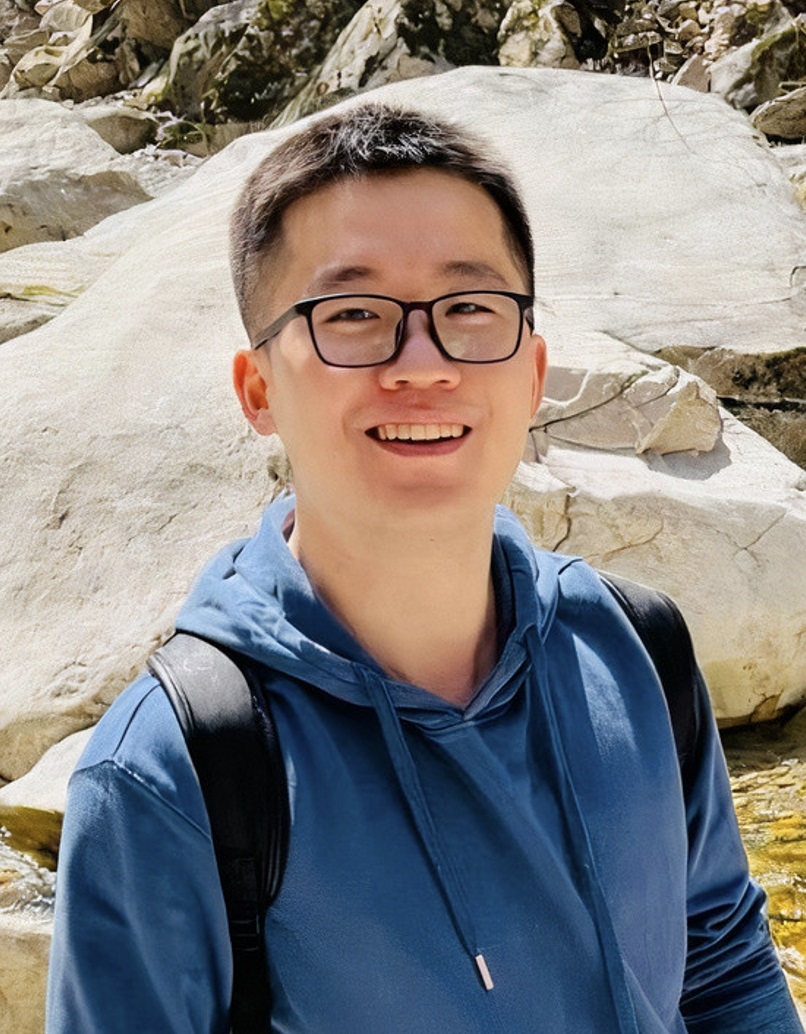}}]{Kaixin Chai}
graduated from Xi'an Jiaotong University in 2022, with a major in Energy and Power Engineering. After graduation, he began his research in robotics and worked as a visiting student at Zhejiang University, City University of Hong Kong, and Korea Advanced Institute of Science \& Technology. His research interests include perception-aware planning, mobile manipulation, and humanoid whole-body control.
\end{IEEEbiography}

\vspace{-2.0cm}
\begin{IEEEbiography}[{\includegraphics[width=1in,height=1.0in,clip,keepaspectratio]{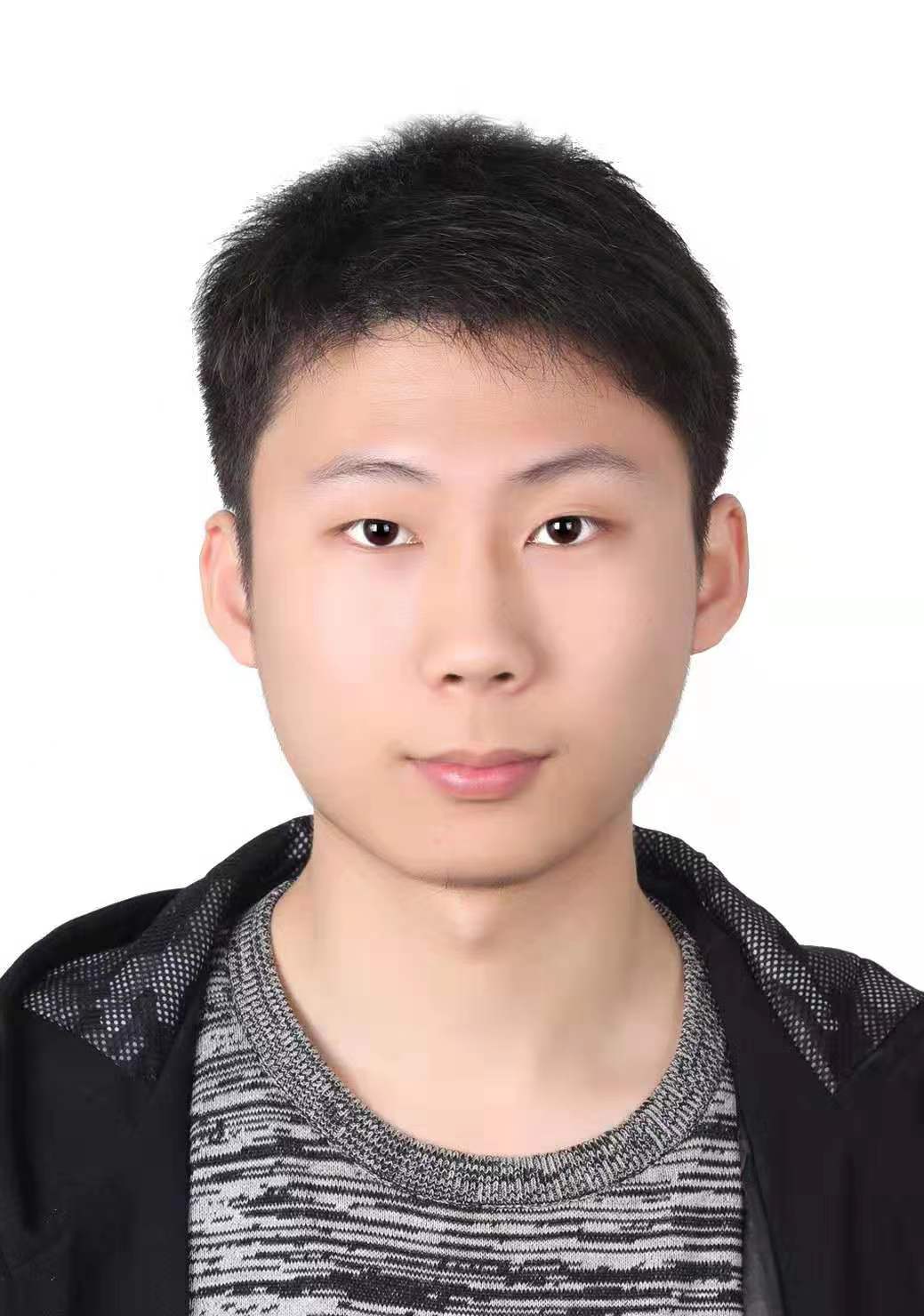}}]{Jialin Ji}
graduated from Zhejiang University, in 2019, with a major in mechatronic engineering at School of Mechanical Engineering and a minor in ACEE (Advanced Honor Class of Engineering Education) at Chu Kochen Honors College. He got the PhD's degree in Automation under the supervision of Fei Gao, at the FAST Lab from Zhejiang University, in 2024. His research interest includes motion planning and autonomous navigation for aerial robotics and unmanned vehicles.
\end{IEEEbiography}

\vspace{-2cm}
\begin{IEEEbiography}[{\includegraphics[width=1in,height=1.0in,clip,keepaspectratio]{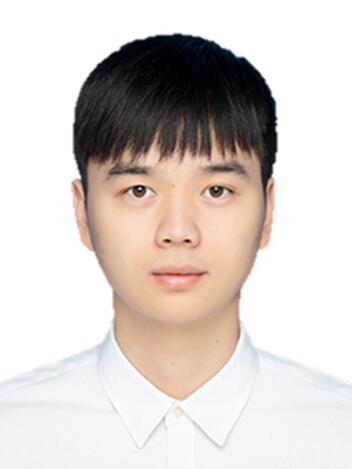}}]{Yuze Wu}
graduated from Zhejiang University in 2021 with a major in Automation from the College of Control Science and Engineering. He is currently pursuing a Ph.D. in Control Science and Engineering at the FAST Lab, Zhejiang University, under the supervision of Professor Fei Gao. His research interests include motion planning and control for aerial robots, reinforcement learning for robotics, and related topics.
\end{IEEEbiography}


\vspace{-2cm}
\begin{IEEEbiography}[{\includegraphics[width=1in,height=1.0in,clip,keepaspectratio]{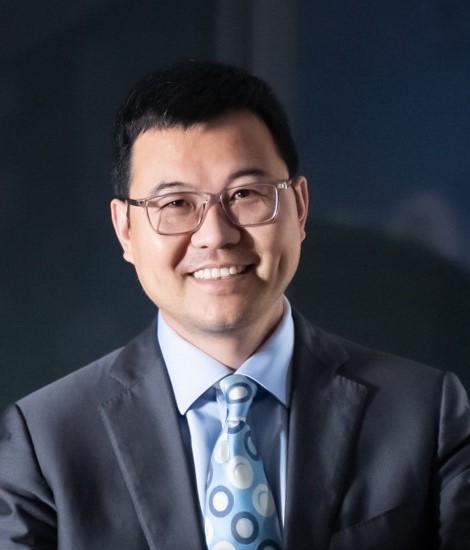}}]{Chao Xu}
received the Ph.D. degree in Mechanical Engineering from Lehigh University in 2010. He is currently Associate Dean and Professor at the College of Control Science and Engineering, Zhejiang University. He serves as the inaugural Dean of ZJU Huzhou Institute, as well as plays the role of the Managing Editor for \textit{IET Cyber-Systems \& Robotics}. 
His research expertise is Flying Robotics, Control-theoretic Learning. Prof. Xu has published over 100 papers in international journals, including \textit{Science Robotics}, \textit{Nature Machine Intelligence}, etc.
\end{IEEEbiography}


\vspace{-1.8cm}
\begin{IEEEbiography}[{\includegraphics[width=1in,height=1.0in,clip,keepaspectratio]{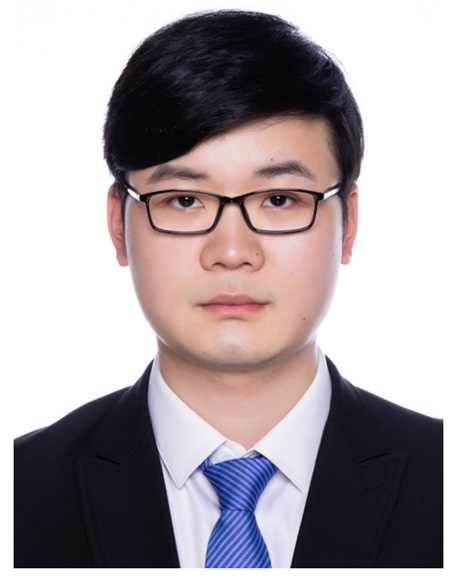}}]{Fei Gao}
received the Ph.D. degree in electronic and computer engineering from the Hong Kong University of Science and Technology,Hong Kong, in 2019. 
He is currently a tenured associate professor at the Department of Control Science and Engineering, Zhejiang University. 
His research interests include aerial robots, autonomous navigation,motion planning, optimization, and localization and mapping. 
He is also the founder of Differential Robotics Ltd. 
\end{IEEEbiography}

\end{document}